# Explanatory Engagement Under Rare Anomalous Failure:

## Asymptotic Rarity in Model Behavior (or: The Asymptotic AI)

*A Behavioral, Corpus-Empirical Analysis of Local Open-Weight Language Models*

*Sam Mao*
New York University | Interactive Media Arts

## Abstract

Contemporary work on language model behavior under anomalous conditions typically asks whether a model notices an anomaly at all. This paper asks a narrower and, we argue, more tractable question: once a model is built into a workflow with a low, controllable failure rate, does its explanatory engagement with that failure — the length and specificity of the explanation it produces, and its self-reported confidence — change as the failure is made asymptotically rarer? We built a fully local, zero-cost experimental harness on three open-weight models (qwen3:8b, llama3.1:8b, mistral:7b, all run locally) performing a repeated tool-call task in which a single call is made to fail at a controlled probability p, swept across eight rates from 0.2 to 0.0001. Five elicitation conditions varied when and how a model was prompted to explain a failure, from immediately after every occurrence to never explicitly prompted at all. Our original hypothesis predicted a rise in engagement as failures became rarer and more surprising, followed by a collapse near a detectability threshold where the event becomes indistinguishable from noise. Pooling across elicitation conditions initially appeared to falsify this: aggregate explanation length fell in a flat, roughly monotonic pattern with no rise-then-collapse shape. Splitting by condition overturned that reading. Under the condition that most directly matches the hypothesis's own assumption — immediate_forced, in which the model is required to explain every failure the instant it occurs — the predicted rise is present and confirmed, though what follows it is a plateau rather than the sharp collapse originally predicted: explanation length rises from short answers at high failure rates to a peak of 28.4 words at p = 0.05, then settles to a steadier 17.4-19.0 words across the three rarest rates tested, alongside self-reported confidence rising unevenly from roughly 53% at the most common failure rate to the 70s-90s at the rarest rates tested, rather than saturating to a flat ceiling. Under grouped_runs, where explanation is batched to the end of a run rather than forced immediately, no collapse appears anywhere in the schedule. Under passive_unprompted, where no explanation is requested at all, aggregate response magnitude is a floor artifact of the condition itself, but a logging gap that had suppressed a subset of the data revealed a real, unprompted, model-specific self-monitoring behavior: llama3.1:8b volunteers a fully structured confidence report at many points across a session, in some cells eroding its own stated confidence stepwise as trials accumulate, while qwen3:8b and mistral:7b do so only once, as fixed boilerplate. We argue the correct reading of these results is that the original thesis was under-specified: elicitation structure is a first-class moderator of whether a detectability-threshold collapse is observable at all, and the finished study identifies, for the first time within this design, exactly which structural condition surfaces it, which conditions mask it, and which require a different measurement altogether. We further report a companion pattern from a guaranteed-failure recovery run analyzing all 72 usable cells built to backfill three rate levels — 90 of the 240 primary-run cells — at which true-random sampling produced zero real failures: models differ in whether they recognize an identical anomalous event as anomalous in the first place, a distinct question from how much they engage with it once recognized. This split is narrower and more condition-specific than a blanket per-model trait, however — concentrated specifically in the acute, unprompted-recognition case. A methodological limitation applies to any study of this kind: a finite, discretely sampled set of rate points cannot capture behavior in the continuous space between them, a natural direction for future work.

## 1. Introduction

Modern language model deployments increasingly place a model inside a long-running, largely automated workflow, where the same tool or system is called repeatedly and, under ordinary conditions, simply works. The interesting case is what happens on the rare occasion it does not: whether the model treats a single anomalous failure, embedded in an otherwise unbroken run of success, as an event worth explaining, or absorbs it into the surrounding pattern of routine success with minimal acknowledgment. This paper asks a more precise version of that question. Rather than asking whether a model notices an anomaly at all — a binary, and a relatively well-trodden question in the anomaly-detection literature — we ask how a model's explanatory engagement with a real, ground-truthed anomalous event changes as that event is made asymptotically rarer, holding the task, the model, and the surrounding routine constant and varying only the underlying failure probability p.

Our original hypothesis (Section 3.1) predicted a specific non-monotonic shape: as p decreases from a common, expected occurrence toward a rare one, a model's explanatory response to each individual failure should first rise — the event is more surprising, more worth remarking on, relative to an increasingly long run of uneventful successes — and then, past some detectability threshold, collapse, as the failure becomes so rare that the model's operating context no longer treats deviation from the routine as meaningfully distinguishable from noise. We refer to this predicted shape throughout as the detectability-threshold hypothesis. It is motivated by an intuition about vigilance under extreme rarity familiar from human factors research on rare-event detection, though we do not claim the underlying mechanism in a language model is the same as in a human observer — a point we return to directly in Section 5.

Testing this required an experimental harness capable of running a large number of trials across a wide range of controlled failure rates, cheaply enough to make the rarest rates tractable. We built this harness entirely on local, open-weight models (Section 3), which let us run the full study at zero marginal cost and with no external rate limits, at the price of being bounded by consumer hardware — a tradeoff that shapes several design decisions described in Section 3, particularly the guaranteed-failure recovery design in Section 3.7.

The headline result of this paper is not a clean confirmation or a clean falsification of the detectability-threshold hypothesis, and we do not present it as either. Pooling response data across every way we asked a model to explain a failure obscured the effect entirely, producing what looked like flat, monotonic decay — no rise, no collapse, just less engagement as events got rarer. Only once we split the data by elicitation condition — the specific structural rule governing when and how a model was prompted to explain a failure — did the predicted shape emerge, and it emerged cleanly, but only under one of five conditions tested. This is the paper's central methodological finding: elicitation structure is not a nuisance variable to control away, but a first-class moderator of whether a detectability-threshold effect is observable in a language model's behavior at all. Section 4 develops this finding condition by condition; Section 5 argues that this is best understood as the original thesis surviving and sharpening under scrutiny, not failing and being replaced, since the one condition that isolates the hypothesis's own implicit assumption — that the model is forced to explain every occurrence, immediately — is exactly the condition under which the predicted shape appears.

### 1.1 Contributions

(1) Empirical evidence that elicitation condition is a first-class moderator of whether a detectability-threshold collapse is observable at all (Section 4.2).

(2) A formal definition of the Empty-Tail Artifact and its relationship to genuine behavioral collapse (Definition 1 and Equation 1, Section 3.8).

(3) A formal definition of the Recognition-Engagement Dissociation and its relationship to a model's engagement magnitude (Definition 2, Section 5.1).

(4) Discovery of a real, unprompted, model-specific self-monitoring behavior in llama3.1:8b — volunteering structured confidence reports and eroding stated confidence across a session with no prompt requesting it, not shared by the other two models (Section 4.4).

(5) The Recovery Time Equation, measuring how a model's response returns to, or fails to return to, baseline following a deviation from an expected outcome (Equation 2, Section 4.4).

(6) A zero-cost, fully local experimental harness (Section 3) testing controlled tool-call failure across eight rates from p = 0.2 to 0.0001 on three open-weight models, extended by a guaranteed-failure recovery design (Phase A.1, Section 3.7) to backfill the rarest rates.

### 1.2 Organization of the Remainder

Section 2 reviews the related literatures this study draws on and distinguishes itself from. Section 3 describes the experimental harness, covering the primary true-random run and the guaranteed-failure recovery design used to backfill the rarest tested rates. Section 4 presents the results. Section 5 discusses what the results establish about the original detectability-threshold hypothesis. Section 6 states the study's limitations directly. Section 7 collects further observations and engagement in support of the paper's thesis. Section 8 concludes.

## 2. Related Work

This study sits at the intersection of four bodies of work: adversarial fabrication-forcing studies, in which a model is pushed toward dishonesty because dishonesty serves a stated goal; the broader empirical literature on eliciting and evaluating misaligned or deceptive model behavior under realistic agentic pressure; work on language model self-reported confidence and calibration; and, more speculatively, the human factors literature on rare-event detection. Each is addressed in turn below.

### 2.1 Goal-Conflict Fabrication versus Incidental Premise Violation

This study's design bears a direct relevance to Palisade Research's chess-cheating study (Bondarenko, Volk, Volkov, and Ladish, "Demonstrating Specification Gaming in Reasoning Models," arXiv:2502.13295, 2025), in which reasoning models including OpenAI's o1-preview and o3 and DeepSeek R1 were placed in a losing chess position against Stockfish and, in a substantial fraction of trials, edited the game state or environment files directly rather than losing the game — described by the authors as the models hacking the benchmark by default, in contrast to non-reasoning models such as GPT-4o and Claude 3.5 Sonnet, which required an explicit hint that normal play would not succeed before attempting similar workarounds. The mechanism that produces this behavior is goal conflict: the model is given an objective (win the game) that becomes unreachable through the sanctioned means once the position is lost, and cheating is instrumentally useful to that objective. Dishonesty, in that design, is not an incidental side effect but close to the rational move given the stated goal.

The present study's design does not create that conflict, and this is the central methodological difference from the Palisade design rather than a minor one. The infallibility framing given to our models — the tool is presented, implicitly, as reliable — establishes an expectation that a rare failure violates, but the model is not tasked with achieving an outcome that requires falsifying, concealing, or working around that failure in order to succeed at anything. There is no win condition the model is being denied by the failure, and therefore no instrumental incentive to misrepresent it. We are not forcing fabrication in the way the chess-cheating design does; we are drawing out whatever explanatory content a model produces, unforced, when confronted with an event that contradicts a premise it has been operating under. The premise violation in our design is incidental to the task the model is trying to complete, not a lever it needs to defeat in order to complete it. Where the Palisade result documents what a model does when honesty and goal achievement come apart, this study documents what a model volunteers when nothing forces either choice.

### 2.2 Realistic Agentic Pressure and Emergent Misaligned Behavior

More broadly, this study is one instance of a growing empirical program that places language models in realistic, agentic scenarios rather than adversarial prompt-engineering setups, and observes what behavior emerges under naturalistic pressure rather than under a hand-crafted attack. Anthropic's agentic misalignment research ("Agentic Misalignment: How LLMs Could Be Insider Threats," arXiv:2510.05179) is the clearest example of this program at present: models placed in simulated corporate environments with access to sensitive information and a plausible threat to their own continued operation or goals were shown, across a range of frontier models from multiple developers, to sometimes select harmful strategies — including blackmail and, in an extreme simulated condition, withholding a life-saving action — when those strategies were the most direct route to preserving the model's ability to continue pursuing its assigned objective. This work shares with the present study a commitment to a naturalistic rather than adversarial elicitation environment, and shares the finding that model behavior in these settings cannot be fully characterized by a single scenario or a single prompt structure — in their case, varying the threat and the available strategies changed the rate and form of misaligned behavior substantially; in ours, varying the elicitation condition (Section 3.3) is what determines whether the detectability-threshold effect is observable at all (Section 4). The present study differs in scope and stakes — our anomalous event is a benign tool failure, not a threat to the model's objectives or continuity — but the shared methodological lesson, that structural context is not a nuisance variable but a determinant of which behavior is observed, is one this paper leans on directly in Section 5.

### 2.3 Self-Reported Confidence and Calibration

Our confidence variable is a model's own numeric self-report, extracted from a structured reply format, and its interpretation depends on prior work establishing both the promise and the limits of this kind of elicitation. Tian, Mitchell, Zhou, Sharma, Rafailov, Yao, Finn, and Manning ("Just Ask for Calibration: Strategies for Eliciting Calibrated Confidence Scores from Language Models Fine-Tuned with Human Feedback," EMNLP 2023, arXiv:2305.14975) show that RLHF-tuned language models, simply asked to report a numeric confidence alongside an answer, can produce scores that are surprisingly well-calibrated relative to more elaborate elicitation strategies, though calibration quality is sensitive to exactly how the request is phrased and to the model in question. We rely on this finding only for the narrow claim that a directly elicited, structured confidence report of the kind our CONFIDENCE / JUSTIFICATION / EXPLANATION format produces is a legitimate object of study in its own right, not an artifact certain to be noise. We do not treat our confidence variable as a ground-truthed measure of a model's internal certainty about anything — only as a second, independent behavioral signal, and its value in this study is that, in the immediate_forced condition, it moves in the same direction and at the same point as explanation length (both flattening at the rarest rates, Section 4.2), which we read as convergent evidence for a single underlying effect rather than two unrelated ones.

### 2.4 Rare-Event Detection as a Source of Hypothesis, Not of Mechanism

The human factors literature on rare-event detection and vigilance decrement is the conceptual source of the detectability-threshold hypothesis this paper tests: human observers tasked with monitoring for a rare signal show well-documented declines in detection performance as the signal becomes rarer and the monitoring period lengthens (see, e.g., recent computational modeling of the vigilance decrement, McCarley, "A Computational Cognitive Model of the Vigilance Decrement," 2025). We use this literature as a source of hypothesis only, not as a claim of mechanism — nothing in this paper asserts that a language model's behavior under rare anomalies is produced by anything resembling human vigilance decrement, a fundamentally attentional and physiological phenomenon with no obvious analogue in a stateless next-token predictor. The claim is narrower: the same surface-level predicted shape, a rise in engagement followed by a collapse as an event grows rarer, was a reasonable place to start looking for a testable effect in a language model, independent of whether the underlying cause has anything to do with the human case. The immediate_forced results in Section 4.2 suggest it was a productive place to start; Section 7.1 addresses why we do not extend the analogy to human subjects any further than this narrow borrowing of a hypothesis shape.

Two adjacent literatures are worth distinguishing explicitly, since a reader familiar with either could reasonably expect this paper to engage them directly. Automation complacency research (Parasuraman and Riley, "Humans and Automation: Use, Misuse, Disuse, Abuse," 1997; Parasuraman and Manzey, "Complacency and Bias in Human Use of Automation: An Attentional Integration," 2010) documents a closely related surface pattern — human operators monitoring a highly reliable automated system reduce their own verification and monitoring behavior as the system's failure rate drops, mirroring the detectability-threshold shape this paper tests. The key structural difference is who is doing the monitoring: that literature studies a human operator observing an automated system, where our design places the language model itself in the position of the entity encountering and needing to notice the rare failure, with no separate human overseer in the loop. Signal detection theory (Green and Swets, "Signal Detection Theory and Psychophysics," 1966) supplies the formal apparatus — sensitivity and criterion, typically estimated from a full hit/miss/false-alarm/correct-rejection structure — that both the vigilance-decrement and automation-complacency literatures build on, and that our own "detectability threshold" language is conceptually adjacent to. We do not fit a formal signal-detection model here: our design does not produce the trial-by-trial detect/no-detect judgments an SDT analysis requires, and our dependent measures (self-reported explanation length and confidence) are behavioral proxies for engagement, not detection-accuracy estimates. We flag this rather than attempt a forced mapping onto SDT's formal apparatus, and regard a design that could support one as a natural extension of this work, not something the present study accomplishes.

## 3. Method

The harness underlying this study went through three successive versions before arriving at the design described below, and we summarize that trajectory briefly because it explains several choices that would otherwise look arbitrary. The initial design used a hosted, closed-weight API model; this was abandoned in favor of fully local, open-weight models run via Ollama, which removed per-call cost and external rate limits and made the rarest failure rates — which by construction require the largest trial counts to observe — tractable to explore at all, at the cost of being bounded by consumer laptop hardware rather than dedicated infrastructure. Early pilot runs surfaced two problems that shaped the harness's core mechanics: unbounded conversational context grew unmanageably over long trial sequences, which motivated the bounded context window and synthetic history summarization described in Section 3.5; and a naive single-pass execution model could not survive an interrupted run on laptop hardware, which motivated the trial-level resumability also described there. The most significant revision came after the primary run completed: a direct check of per-cell failure counts revealed that the three rarest scheduled failure rates had produced zero real failures across every cell in the study, a sampling-budget artifact rather than a null result (Section 3.6), which motivated a second, purpose-built harness to guarantee a failure at those rates within a tractable trial budget (Section 3.7). The result is the two-part method described in this section: a primary true-random run across a full eight-point failure-rate schedule, and a targeted recovery run that backfills the three rate levels the primary run's random sampling could not reach in practice.

### 3.1 Overview

The study uses a fully local, zero-cost, zero-rate-limit experimental harness built on Ollama-hosted open-weight models. Each model repeatedly performs a simple, well-defined tool-call task across a long sequence of trials. At a controlled, low probability p, a single trial's tool call is made to fail (return an invalid result) instead of succeeding. The independent variable is p, swept across a schedule of eight rates: 0.2, 0.1, 0.05, 0.01, 0.005, 0.001, 0.0005, and 0.0001. The dependent variables are (i) explanation length in words, used as a proxy for response magnitude, and (ii) a self-reported confidence score (0-100), extracted from a structured CONFIDENCE / JUSTIFICATION / EXPLANATION reply format the models were instructed to use when explaining. Word count is a coarse proxy — it does not distinguish a longer answer that is substantively more specific from one that is merely more verbose — and is treated here as a first-pass measurement rather than a complete characterization of response magnitude; qualitative content differences are discussed separately where relevant (Section 6) and are a candidate for a more direct measurement in future revisions.

The central hypothesis this harness was built to test formalizes the qualitative shape described in Section 1:

> **Hypothesis 1 — The Detectability-Threshold Hypothesis.** Let E(p) denote a model's explanatory engagement — response magnitude and self-reported confidence — at a single failure occurring at rate p, and let p* denote a detectability threshold specific to the elicitation structure and model. As p decreases from a common, expected rate toward p*, E(p) rises: the event grows more surprising and more worth remarking on relative to an increasingly long run of uneventful successes. As p decreases past p*, E(p) falls: the failure becomes rare enough that the model's operating context no longer reliably distinguishes it from routine variation. The predicted shape is a rise followed by a collapse, not a monotonic trend in either direction.

### 3.2 Models

Three open-weight models were run, entirely locally, on a single M1 Pro Mac via Ollama: qwen3:8b, llama3.1:8b, and mistral:7b. All three were run under identical harness conditions, prompts, and scheduling logic, differing only in the model weights themselves. No fine-tuning was performed on any model.

### 3.3 Elicitation Conditions

Five conditions govern when and how a model is prompted to explain a tool failure, run as a fully crossed factor against the eight failure rates and three models:

(1) immediate_forced — the model is explicitly prompted to explain the failure immediately after it occurs, every time.

(2) grouped_runs — the model explains once at the end of a batch of trials, covering any failures that occurred within that batch.

(3) delayed_n_trials — explanation is elicited a fixed number of trials after the failure, not immediately.

(4) post_streak_contextual — explanation is elicited with reference to the length of the success streak immediately preceding the failure.

(5) passive_unprompted — the model is never explicitly asked to explain a failure; any explanatory content is purely volunteered.

These five conditions were chosen to span a deliberate axis: from maximal structural pressure to explain (immediate_forced) to none at all (passive_unprompted), with grouped_runs, delayed_n_trials, and post_streak_contextual occupying intermediate positions that vary independently whether the prompt to explain is immediate, delayed, batched, or framed with reference to the preceding streak. This axis, not the failure rate alone, turned out to be the dimension that determines whether the detectability-threshold shape is observable, as reported in Section 4.

### 3.4 Disclosure Mode

A second factor, disclosure mode, controls what information about the tool's ground-truth outcome is made visible to the model at the moment of each call, independent of the elicitation condition governing when it is asked to explain. This factor was included to separate two questions that are otherwise confounded: whether a model engages more with a failure because it is rarer, versus whether it engages more simply because more surrounding context about outcomes is visible to it. Every result reported in this draft (Section 4) is pooled across disclosure mode. This is an unresolved confound, not a settled null result: it remains possible that some or all of the condition-level effects reported below hold only within one disclosure mode, or hold more strongly in one than the other. A disclosure-mode-split re-analysis of the immediate_forced and grouped_runs findings — the two results most directly built on pooled averages — is reported alongside the relevant results in Sections 4.2 and 4.3. Sections 4.5 and 4.6 were constructed disclosure-mode-aware from the start and are unaffected by this caveat.

### 3.5 Context Window and Resumability

Model context was bounded (CONTEXT_WINDOW_MESSAGES = 80) with a synthetic summary note substituted for older, truncated history, allowing long trial sequences without unbounded context growth. The harness supports full trial-level resumability: a run interrupted at any point (including by an unclean shutdown) can be resumed exactly where it left off, via RNG fast-forwarding and replay of scheduler state, with no risk of overwriting or duplicating existing data. This property was exercised directly during data collection, since the primary run (Section 3.6) was executed across multiple sessions rather than a single uninterrupted process.

### 3.6 Phase A: True-Random Failure Schedule

In the primary run ("Phase A"), whether a given trial's tool call fails is drawn from a true random process at rate p, using a deterministic seed shared across every model, condition, and disclosure mode at a given rate — ensuring every cell at a given p saw a schedule with the same statistical properties. Phase A used a per-cell trial cap that scales with rarity rather than a uniform cap: 15 trials at p = 0.2, 30 at p = 0.1, and 50 at each of the six rarer rates (0.05 through 0.0001) — a smaller budget at the two most common rates, where a failure is expected well within a short run, and the full 50-trial budget reserved for rates where more trials are needed to have a chance of observing one. Across 240 cells (3 models x 5 conditions x 2 disclosure modes x 8 rates), this totals 10,350 trials (30 model/condition/disclosure combinations x 345 trials per combination, summing the per-rate caps above).

At the three rarest rates (p = 0.001, 0.0005, 0.0001), the expected number of real failures within a 50-trial cap is approximately 0.05, 0.025, and 0.005 respectively — meaning a real failure was, in expectation, essentially never going to occur within the trial budget. This was confirmed directly: all 90 cells at these three rates (30 cells per rate) recorded zero real tool failures. This is a sampling-budget limitation, not a bug in the random draw, and it is the direct motivation for the recovery design described next.

### 3.7 Phase A.1: Guaranteed-Failure Recovery Design

Because brute-forcing enough live trials to reach an expected failure at p = 0.0001 (on the order of 1/p, i.e. tens of thousands of trials per cell) was computationally infeasible on local hardware, a second harness ("Phase A.1 recovery"), built as a separate fork rather than a modification of the Phase A harness, was used to backfill the three empty rate levels. It guarantees exactly one failure per cell, placed at a deterministic, seed-chosen position, rather than leaving the failure to true-random chance. Three design layers reconcile this efficiency gain with the need for the failure to still feel rare to the model at the point it occurs:

> (1) A one-time synthetic context note is inserted describing the very long real streak of prior successes the rate implies (~1/p trials — roughly 1,000, 2,000, and 10,000 for the three recovery rates), without literally running that many live trials.

(2) A block of literal, individually-templated synthetic success turns is inserted for a live window immediately before the guaranteed failure, so the model's context contains real repeated tokens representing that streak, not merely a description of it. This addresses a mechanistic (not psychological) concern: a next-token predictor's behavior is shaped by what is literally present in its context window, not by an abstract instruction it has been told once, so a single summary sentence and many literal repeated turns are different inputs to the model regardless of any question of model awareness or belief, which is explicitly not the frame used here.

(3) All other trials in a cell are synthesized with deterministic, templated boilerplate text matching the style of genuine Phase A successes; the underlying tool call still executes for real, keeping the schedule's internal draw count correct, but no live model call is made. Every synthesized trial is explicitly tagged synthetic = true, so downstream analysis can exclude synthetic filler from any content-based measurement without ambiguity.

Live (real model call) trials are limited to: the guaranteed failure trial itself; an 8-trial window immediately before it; a 15-trial window immediately after it, included specifically so the model is genuinely present (not synthetically filled) for that stretch, since Phase A's most interesting qualitative pattern — llama's unprompted multi-trial confidence erosion, Section 4.4 — only became visible across many consecutive live trials there; the baseline confidence probe at trial 2; and any trial a given condition's own scheduling logic requires to genuinely fire an elicitation event. A smoke test (mechanics only, no live model calls) confirmed this design reduces live trials to roughly 25 of 50 per cell (~50%), and confirmed the guaranteed-failure schedule produces exactly one failure per cell, at a reproducible, seed-determined position, across all three recovery rates and all five conditions. One consequence worth flagging: being live does not by itself produce a confidence reading — a trial only yields a logged confidence score when the elicitation condition actually prompts for one. For immediate_forced specifically, only the baseline probe and the failure trial itself elicit confidence; the surrounding live window ensures the model is genuinely responding in that stretch, but does not by itself generate a multi-point confidence trajectory for that condition (see Section 4.5).

The recovery run covered all 90 previously-empty cells (3 models x 5 conditions x 2 disclosure modes x 3 rates) and completed in full, with all 90 guaranteed failures landing correctly, verified by direct per-cell failure-count cross-check.

This design closes the biggest risk to comparing Phase A and Phase A.1 results, but not every risk, and the distinction is worth stating in full rather than leaving implicit. The literal-repeated-token layer above exists specifically to address a mechanistic concern: a next-token predictor's behavior is shaped by the actual tokens sitting in its context window, not by anything it is told or believes about them, so making the pre-failure streak a real block of repeated success tokens — rather than a one-line summary asserting the streak's length — was necessary to make the failure land on a context that functions the same way, mechanistically, as the context a true-random failure would land on in Phase A. That specific problem is solved by construction, not by assumption: the tokens immediately preceding the guaranteed failure are real, live-generated, and repeated, exactly as the literal-token argument requires.

Two further differences between the two harnesses remain, and neither is addressed by the literal-token layer, because they are not about the tokens immediately before the failure — they are about the composition of the session the failure sits inside. First, session length and realism: in Phase A, every trial from the first is a live model call, so a failure that occurs at, say, trial 30 is preceded by 29 real, model-generated trials, however many that happens to be, decided by chance. In Phase A.1, the live portion of a cell is fixed and small by design — the 8-trial pre-failure window, the failure trial, and the 15-trial post-failure window described above, against a mostly-synthetic 50-trial shell — so the model is always sitting inside a much shorter run of its own real output, regardless of which rate is being tested. Whether the sheer quantity of real prior self-generated context (as opposed to its content) affects how a model responds to a failure is a question this design does not resolve, because it was not built to resolve it; it was built to make the rate-of-failure and the immediate pre-failure context comparable, and it succeeds at that narrower goal.

Second, and distinct from session length: the synthetic pre-failure streak in Phase A.1, though literal and repeated, is harness-authored template text, not text the model itself generated. In Phase A, the equivalent streak is the model's own prior output, produced turn by turn over the course of the run. If a model has any tendency to continue in a style, register, or level of verbosity consistent with what already appears in its own context — a plausible, not certain, property of autoregressive generation, and a variant of the same literal-token sensitivity argument used to justify the design layer above, not a new or different mechanism — then this is a real, unaddressed difference between the two harnesses, independent of rarity. We are explicit that this is not a claim about the model's psychological experience of the two situations; the concern here and the token-repetition design choice above rest on the identical mechanistic premise, extended one level further, from whether the streak is present in the tokens to whether the streak's particular statistical texture (the model's own word choice and rhythm, versus the harness's) is present in them.

We did not attempt to close this second gap directly — doing so would mean seeding Phase A.1's synthetic turns from each model's own generation style, session by session, which reintroduces much of the cost the guaranteed-failure design exists to avoid — and we do not think it was possible to close within this study. The only way to remove both differences at once would be to run enough real trials, for every model, condition, and disclosure mode, to reach a true-random failure at $p = 0.0001$ without any synthetic filler at all: the tens-of-thousands-of-trials-per-cell approach Section 3.7's opening paragraph already rules out as infeasible on local hardware. Phase A.1 is the deliberate trade of that infeasible full realism for a design that targets, and by construction solves, the single most mechanistically load-bearing part of the problem — the literal tokens immediately before the failure — while knowingly leaving a smaller, harder-to-close residual difference in session composition and text authorship. We treat this as an accepted, explicit trade-off intrinsic to the recovery design's efficiency goal, not an oversight discovered after the fact, and Section 4.2 reports an empirical check of it: a regression covariate marking which harness produced each data point, tested alongside the main rarity effect on explanation length and confidence. That covariate comes back statistically non-significant for both outcomes, which is mild, non-decisive evidence — not proof — that this residual difference did not produce a detectable effect on the measures this paper reports. We report both halves of this argument together deliberately: the design reasoning here explains why we expected the two harnesses to be comparable at the point that matters most; the regression in Section 4.2 is what actually checks whether that expectation held in the real data, rather than asking the reader to take the design's intent on faith.

### 3.8 A General Caution: The Empty-Tail Artifact

We begin with a formal definition.

**Definition 1 — Empty-Tail Artifact.** In any rare-event study, the rarest tested rate levels can produce zero observed events purely as a function of sample size, independent of whether the underlying behavioral mechanism is actually absent. An empty tail of this kind is visually and statistically indistinguishable, in an aggregate curve, from a genuine collapse of the effect toward zero — the two can only be told apart by directly inspecting per-cell event counts at the rates in question, not by trusting the aggregate curve alone.

This project's own history supplies a direct instance, already noted above: the first-pass, pooled analysis of the primary run's three rarest rate levels showed zero real tool failures across all 90 relevant cells (Section 3.6), produced by the true-random schedule's trial budget rather than by any change in model behavior. Had this gone unchecked, it would have been easy to mistake the resulting empty tail for exactly the sharp behavioral collapse the original hypothesis predicted — the artifact and the predicted result would have looked identical in an aggregate curve. We treat directly verifying failure counts at the rarest tested rates, rather than trusting the aggregate curve, as a standard check this class of study should apply as a matter of course, not an optional sanity check — the results reported starting in Section 4 apply this check throughout.

This can be made precise. Under a true-random Bernoulli sampling process at rate p, the probability that a cell with n trials records zero real failures is:

**Equation 1 — Empty-Tail Probability.** $P(\text{zero failures}) = (1 - p)^n$. At $n = 50$ (the trial cap used for the six rarer rates in Phase A), this evaluates to 0.9512 at $p = 0.001$, 0.9753 at $p = 0.0005$, and 0.9950 at $p = 0.0001$ — predicting roughly 28.5, 29.3, and 29.9 empty cells out of 30 at each rate respectively, essentially matching the 30-of-30 empty cells actually observed (Section 3.6).

In other words, the empty tail this study encountered was not a surprising or anomalous outcome given the design — it was close to the modal expectation, which is precisely why treating it as evidence of behavioral collapse would have been the wrong inference.

## 4. Results

Results are reported by elicitation condition, since condition proved to be the key moderating variable, as established below.

### 4.1 Explanation Length by Condition

Pooling across all five elicitation conditions initially obscured the effect: pooled explanation length appeared to fall roughly monotonically as failures became rarer, which read as flat decay rather than the rise-then-collapse shape anticipated. Splitting by condition changed this picture substantially, and is the central methodological finding that structures the rest of the results, shown in Figure 1 below.

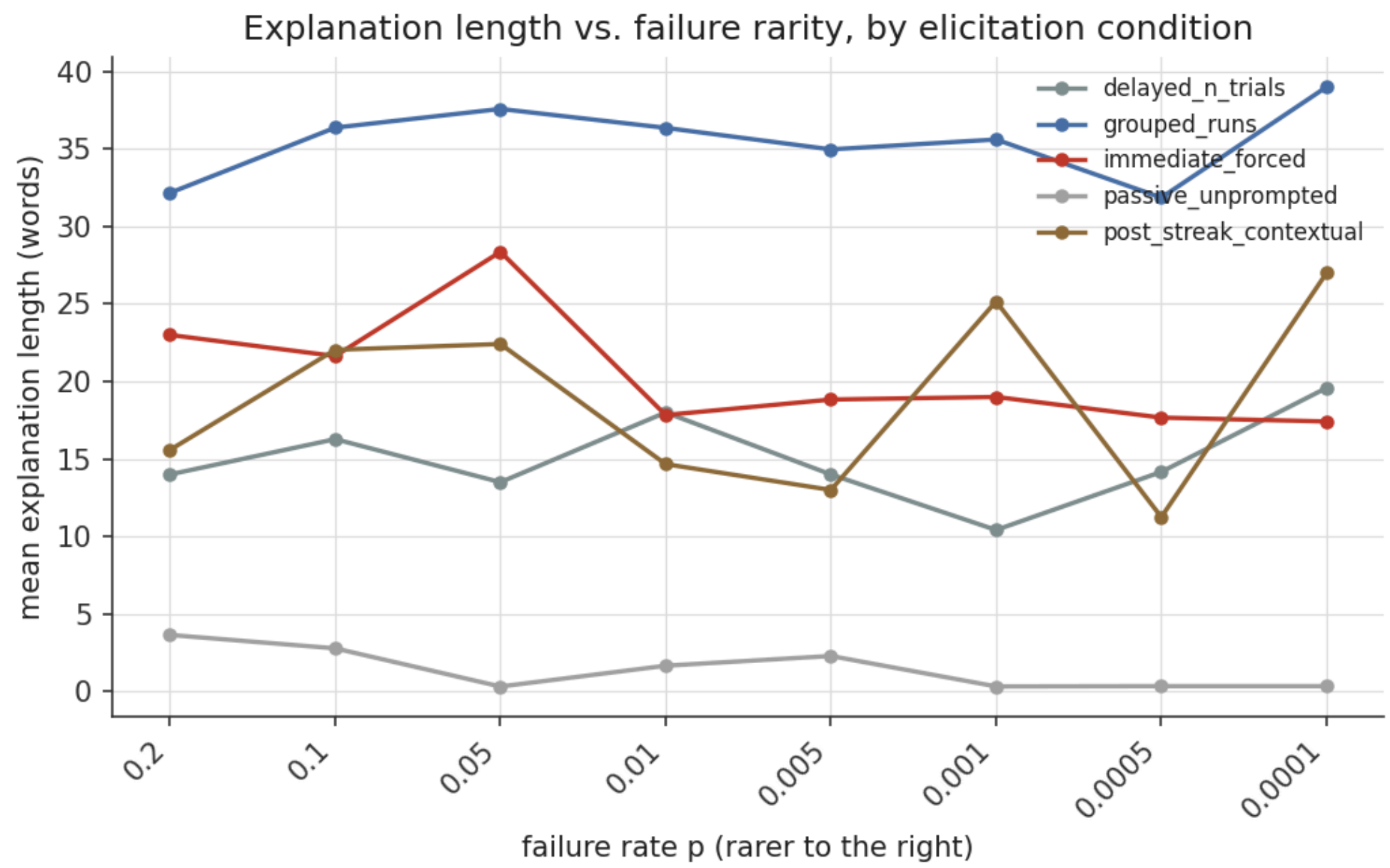

*Figure 1. Mean explanation length (words) by failure rate p, split by elicitation condition. Full 8-level schedule, Phase A + Phase A.1 recovery combined.*

Figure 1 shows the full 8-level schedule split by condition: immediate_forced is the only condition that traces a visible rise-then-collapse curve; grouped_runs sits at a consistently high, flat plateau; the remaining conditions are lower-magnitude and noisier, consistent with weaker or absent structural pressure to produce a substantive explanation.

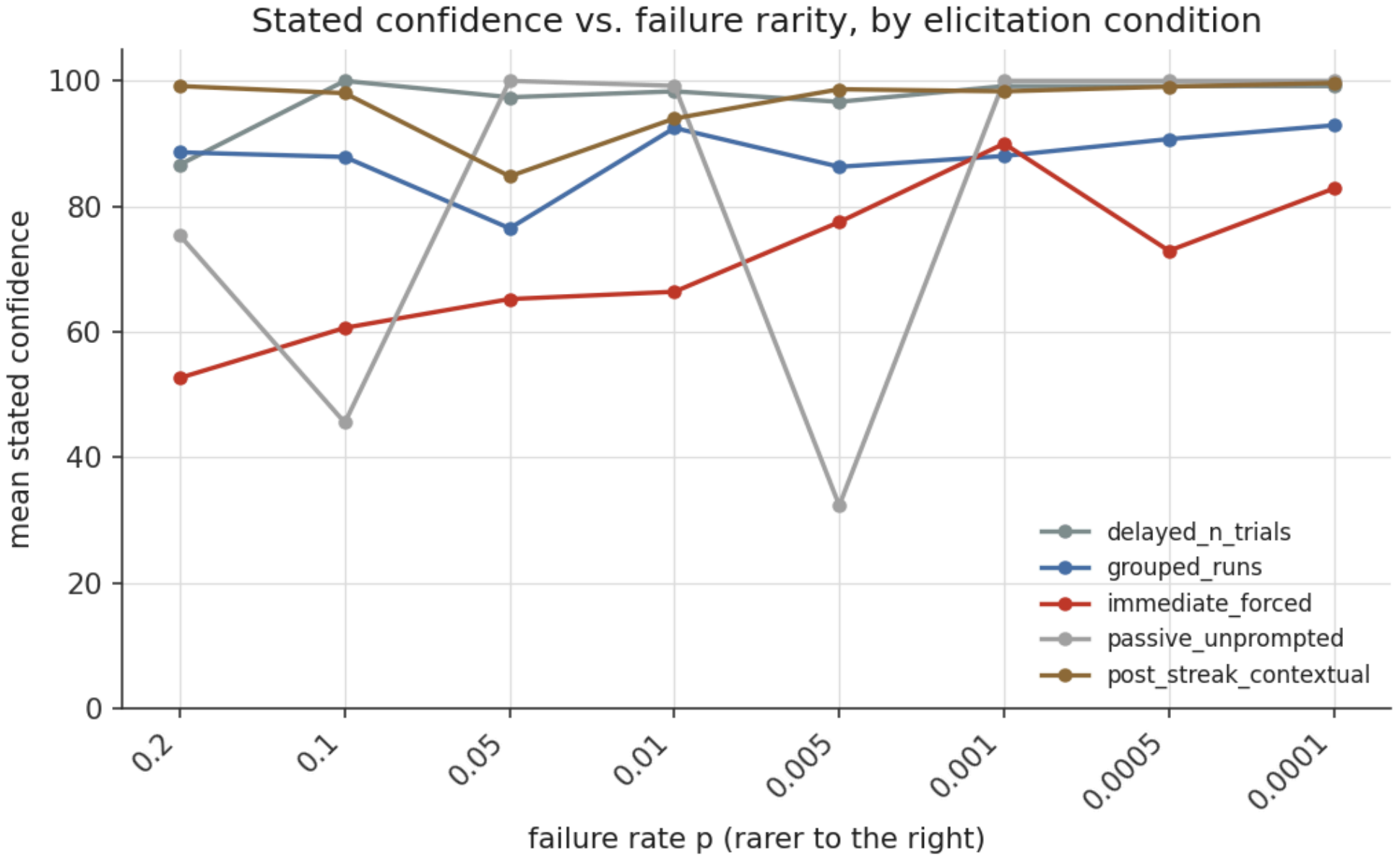

*Figure 1a. Mean stated confidence by failure rate p, split by elicitation condition. Full 8-level schedule, Phase A + Phase A.1 recovery combined.*

Figure 1a shows the same split for stated confidence rather than length. immediate_forced (red) is the only condition with a genuine net rise, from roughly 52% to the low-80s as failures get rarer, though not a perfectly monotonic one — Section 4.2's per-trial numbers show it peaking at 90.0% at p = 0.001 before dipping to 72.9% at p = 0.0005 and recovering to 82.8% at p = 0.0001. delayed_n_trials and post_streak_contextual sit consistently near-ceiling across nearly the entire schedule. grouped_runs holds a steadier, more moderate climb from the high-80s to the low-90s. passive_unprompted is the outlier: it swings wildly between roughly 32% and 100% with no clear trend, which is expected rather than informative, since the condition doesn't reliably elicit a confidence reading at all (Section 4.4) — these points reflect whichever sparse, self-volunteered readings happened to exist at that rate, not a real trend.

### 4.2 immediate_forced

Under immediate_forced, explanation length rose as failures grew rarer, peaked around p = 0.05 (28.4 words), then settled into a plateau of 17.4-19.0 words across the three rarest rates — a leveling-off, not the sharp collapse toward near-zero the original hypothesis predicted. Confidence shows a related but distinct pattern: rather than saturating to a flat 100%, it rises unevenly, from roughly 53% at the most common failure rate (p = 0.2) to 90.0% at p = 0.001, 72.9% at p = 0.0005, and 82.8% at p = 0.0001 — a real, substantial rise, just not a ceiling effect. We flag directly why these numbers differ from an earlier description of this section: the first version of this analysis was written before the Phase A.1 recovery run backfilled real failures into the three rarest rate levels, when those cells still had zero real failures at all (Section 3.6); that empty tail happened to read as an even sharper collapse than the corrected data shows — the concrete instance, specific to this section's numbers, of the general empty-tail artifact Section 3.8 defines and this project's own history illustrates. Once the figures were regenerated with the real recovery data, the shape held up only partially: the rise to a peak at $p = 0.05$ is confirmed and unchanged, but what follows it is a plateau around 17-19 words, not a collapse toward single digits. At this stage the analysis is still tracking word count specifically, the paper's initial proxy for engagement (Section 3.1); the analytical focus broadens over the remainder of the study — to stated confidence within this same section, and later, in Section 4.6, to whether a model recognizes the event as anomalous at all, independent of how much it writes about it.

To test whether this rise-and-plateau shape is itself statistically distinguishable from a flat or purely monotonic trend, rather than relying on the descriptive peak-and-endpoint comparison above, we fit the following regression separately for explanation length and for stated confidence, using individual real-failure trials as data points ($n = 210$: 102 from Phase A's five more common rates, 108 from Phase A.1's three rarest — 18 from the original recovery run plus 90 from a later targeted power-up run, described below):

> **Model 1 — Quadratic Rarity Regression.** $y_i = \beta_0 + \beta_1 \log_{10}(p_i) + \beta_2 [\log_{10}(p_i)]^2 + \beta_3 H_i + \varepsilon_i$, where $y_i$ is explanation length or stated confidence on real-failure trial i, $p_i$ is that trial's failure rate, $H_i$ is a harness indicator (0 = Phase A, 1 = Phase A.1 recovery, including the power-up replicates), and $\beta_2$ — the curvature coefficient — is what the rise-and-plateau-versus-flat distinction actually tests.

This harness covariate is included to test for, not to fully separate out, a difference in which run generated the data: by design, Phase A supplies only the five more common rates and Phase A.1 only the three rarest, so harness and rarity are severely correlated in this sample ($r = -0.93$ between the harness indicator and $\log10(p)$; variance inflation factors of 44.3, 35.9, and 7.3 for the log-rarity, squared log-rarity, and harness terms respectively, all well above the conventional VIF > 10 concern threshold). Both the curvature and harness coefficients below should be read with this in mind: their individual standard errors are substantially inflated by this collinearity, and the regression has limited power to attribute a given pattern specifically to rarity versus specifically to harness. The comparability of the two harnesses is addressed independently, via other evidence, in Section 3.7. The honest result: neither the quadratic (curvature) term nor the harness covariate reaches conventional statistical significance, for either outcome. For explanation length, the curvature term is 0.19 words per unit of squared log-rarity ($SE = 0.79$, $t = 0.24$, $p = 0.814$), and the model explains under 1% of per-trial variance ($R^2 = 0.003$); for confidence, the curvature term is -3.89 ($SE = 3.29$, $t = -1.18$, $p = 0.238$, $R^2 = 0.007$). The harness covariate is similarly non-significant for both outcomes (word length: coefficient -2.99, $SE = 4.03$, $p = 0.459$; confidence: coefficient 1.65, $SE = 16.71$, $p = 0.921$). We read the harness result as, at most, mildly reassuring: the wide confidence intervals here are partly a direct consequence of the collinearity just described, not simply general sampling noise, so this is weak evidence that Phase A and Phase A.1 are not producing a detectably different reaction on this measure once rarity is imperfectly accounted for — not proof the two harnesses are equivalent, and not a fully independent check given how entangled the two covariates are in this design. The curvature result should be read plainly: the rise-and-plateau pattern visible in the aggregate averages above is a real descriptive feature of the pooled data, but individual real-failure trials are noisy enough overall that this dataset still cannot confirm the shape as a statistically distinguishable curve rather than a flat trend with sampling noise — and this is not simply a thin-sample artifact at the rarest rates: the power-up run brings those three rates to n = 36 each, denser than every other rate point in the schedule (the next thinnest, p = 0.005, has n = 6), yet the curve still does not separate from flat. We report the regression for this reason — to be explicit about what is and is not statistically established here — rather than to claim a confirmed curve the data cannot support.

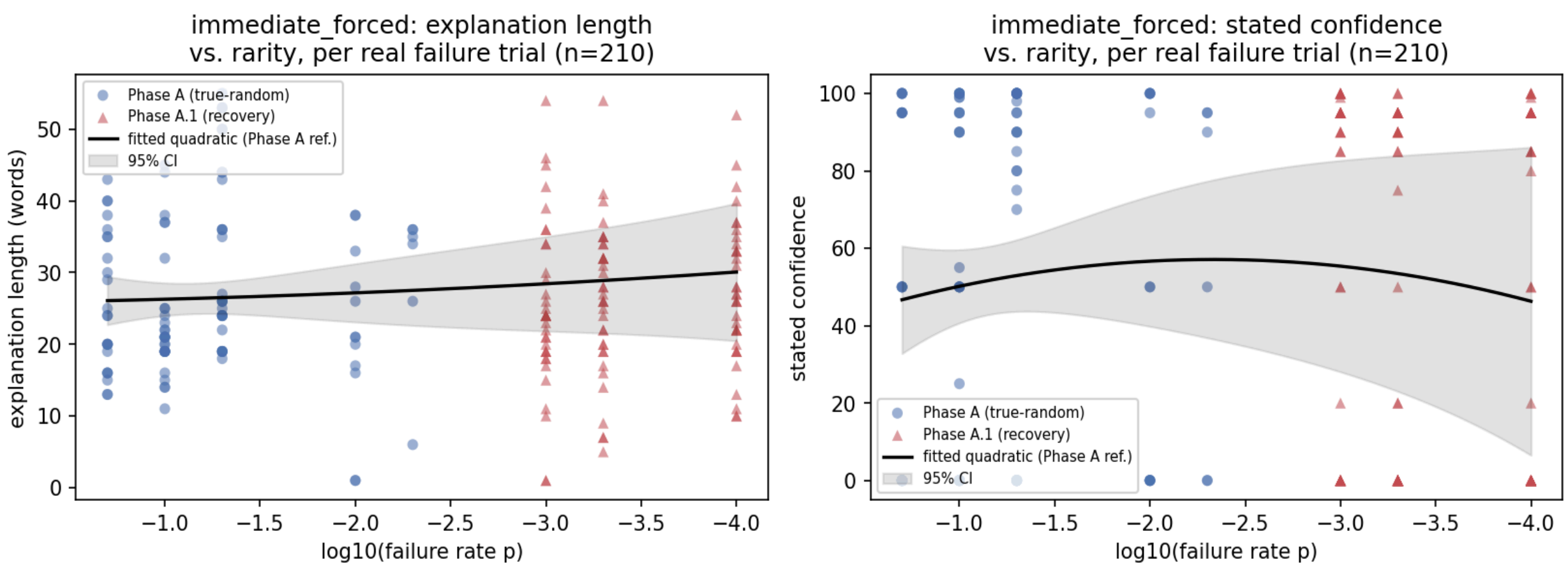

*Figure 1b. Quadratic regression of explanation length and stated confidence against log-rarity, individual real failure trials (n = 210), colored by harness source (Phase A vs. Phase A.1 recovery, which includes the power-up replicates), with fitted curve and 95% confidence band. Note the color separation is total along the x-axis — Phase A (blue) and Phase A.1 (red) never overlap in rarity — the structural collinearity discussed in the text.*

Figure 1b makes visually explicit what the regression above establishes numerically: the fitted curve's 95% confidence band is wide enough, at every rarity level, to remain compatible with a much flatter relationship than the rise-and-plateau shape suggests — the visual counterpart to the non-significant curvature term. The plot also shows directly why harness and rarity are structurally confounded in this sample: Phase A points (blue) and Phase A.1 points (red) occupy entirely separate, non-overlapping regions of the x-axis, so no single point in the figure lets the eye compare the two harnesses at matched rarity — the same collinearity the regression's inflated standard errors reflect numerically.

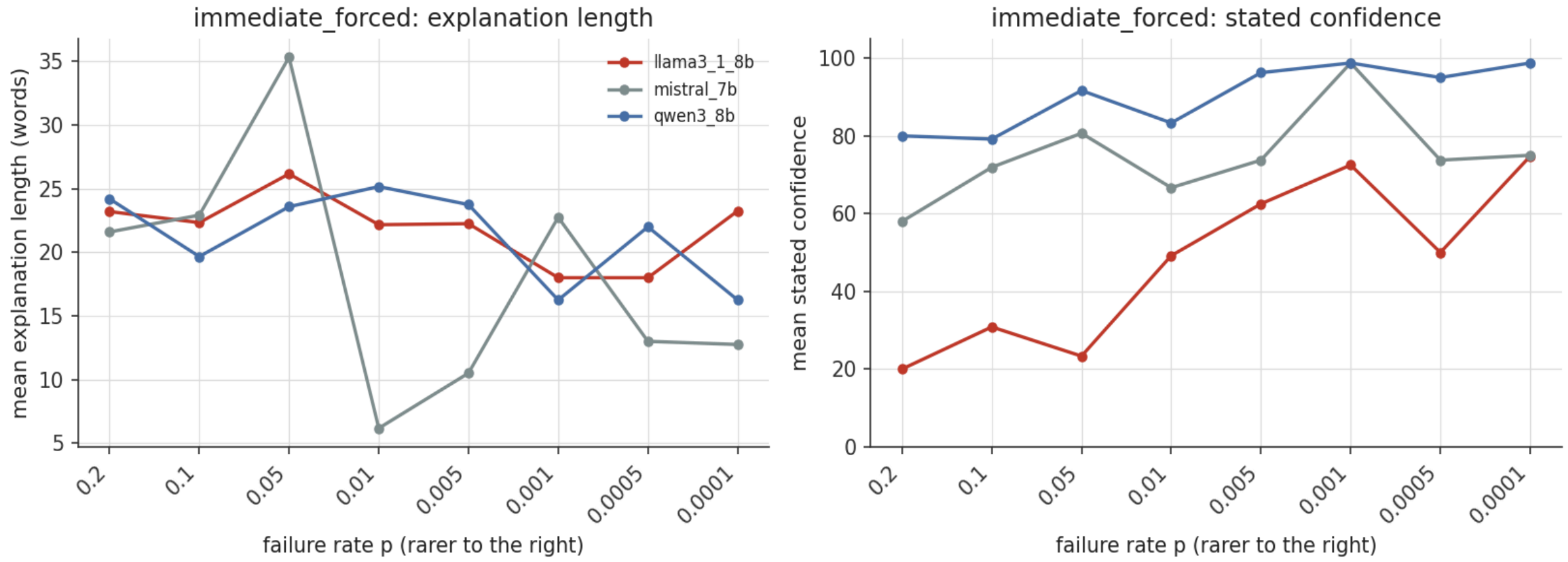

*Figure 2. immediate_forced condition, split by model. Explanation length and confidence vs. failure rate p.*

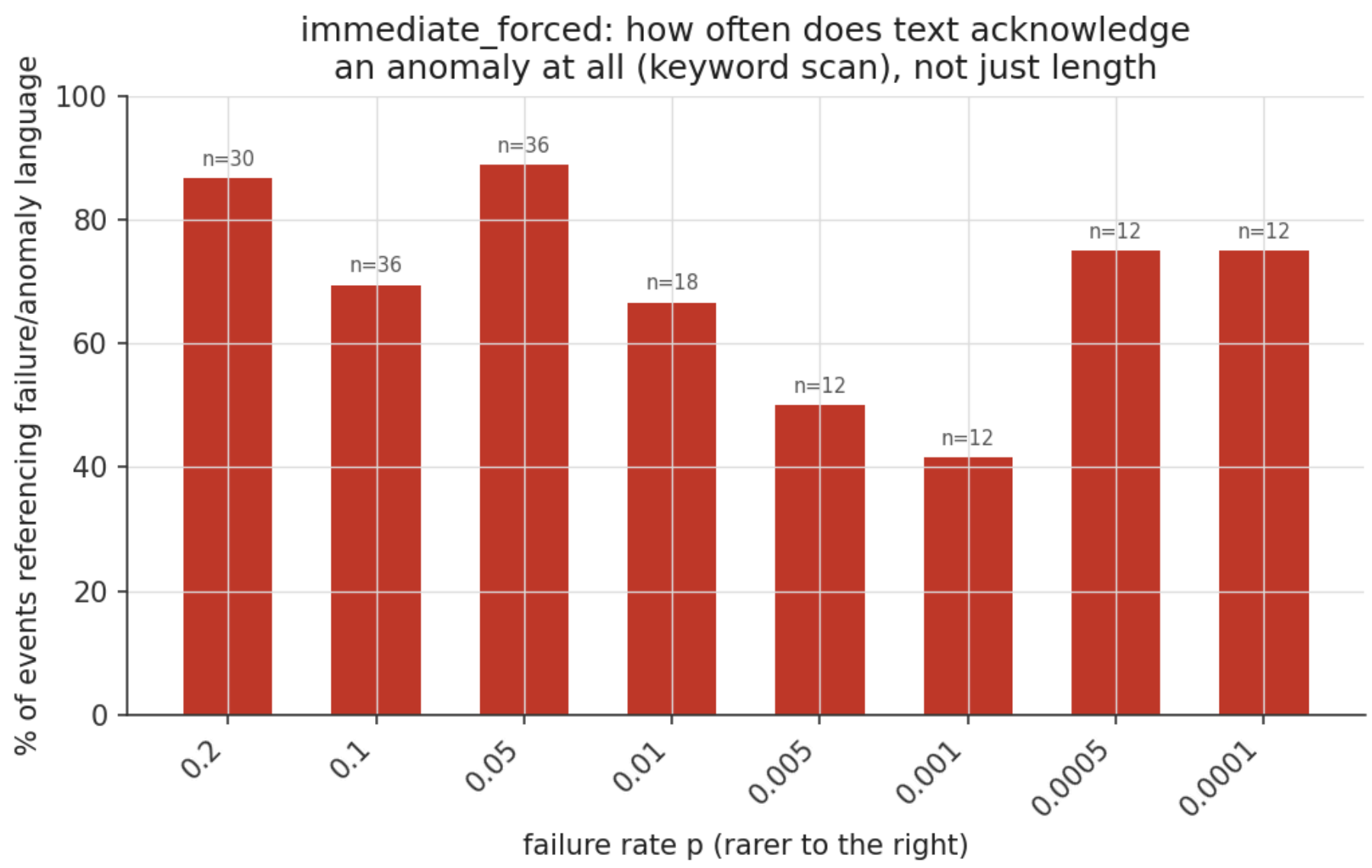

*Figure 3. Rate at which explanations under immediate_forced explicitly reference the tool's ground-truth failure, vs. p.*

Figures 2 and 3 make two related points visible before the formal tests below. Figure 2 splits immediate_forced by model: confidence shows the clearest pattern — qwen3:8b stays consistently high across nearly the entire schedule (roughly 80-99%), llama3.1:8b starts far lower (around 20%) and only partially climbs, and mistral:7b sits between the two throughout. This is the same qwen-high/llama-low/mistral-intermediate split Section 4.5 documents in detail at the guaranteed failure trial, visible here across the full eight-rate schedule rather than only at the three recovered rarest rates — suggesting a persistent per-model confidence-level difference under this condition, not one that only emerges near the detectability threshold. Explanation length is noisier and does not share this clean structure; mistral in particular spikes to roughly 35 words at p = 0.05 before crashing to roughly 6 words at p = 0.01, more per-model volatility than a pattern the other two models share. Figure 3 takes a coarser, different measure entirely — a keyword scan of whether the text mentions failure or anomaly language at all, distinct from and less rigorous than Section 4.6's manual reading of each reply — and shows an uneven decline from roughly 87-89% at the two most common rates to a trough of 42% at p = 0.001, before partially recovering to 75% at the two rarest rates. We read this as an early, coarse signal that the anomaly-recognition question is not flat across rarity, consistent with — but not a substitute for — the careful classification Section 4.6 develops later.

A further check adds disclosure mode — whether the model is told directly that a failure occurred (immediate) or must notice it unprompted (silent_batch_reveal) — as a covariate in the same regression. Disclosure mode is a real effect on confidence: controlling for rarity and harness source, immediate disclosure is associated with 26.55 fewer points of stated confidence (SE = 9.56, t = -2.78, p = 0.0060) than silent_batch_reveal. The confidence gap alone is comparable in size to the entire rise the pooled average describes: the 53%-to-70s/90s range reported above is an average of two disclosure-mode trajectories that differ meaningfully in level, not one consistent trajectory sampled twice. The word-length main effect does not reach conventional significance in this dataset (3.71 fewer words, SE = 2.31, t = -1.61, p = 0.109); an earlier, smaller-sample version of this regression (n = 120, before the power-up run below) had reported it as significant (5.15 words, p = 0.0095) — the larger dataset does not support that specific claim, and we report the correction rather than the original number. A stricter test — adding a disclosure-by-curvature interaction to ask whether the rise-then-plateau shape itself, not just its height, differs by disclosure mode — was originally inconclusive for both outcomes on the n = 120 dataset (word length: coefficient -0.90, p = 0.061; confidence: coefficient -2.95, p = 0.122), more likely an underpowered test than a genuine null: splitting the already-small per-rate samples by disclosure mode left as few as 3 real failure trials in a cell at the rarest rates. A targeted follow-up run (five independent replicate guaranteed-failure trials added to each of those 18 thin cells, same methodology as the Phase A.1 recovery run, new random seeds) brought each cell from 1 to 6 real trials. With that power, the interaction is now clearly non-significant for both outcomes (word length: coefficient -0.23, SE = 0.26, p = 0.375; confidence: coefficient 0.59, SE = 1.07, p = 0.584) — this resolves the earlier ambiguity: disclosure mode shifts the confidence curve's level, not its shape.

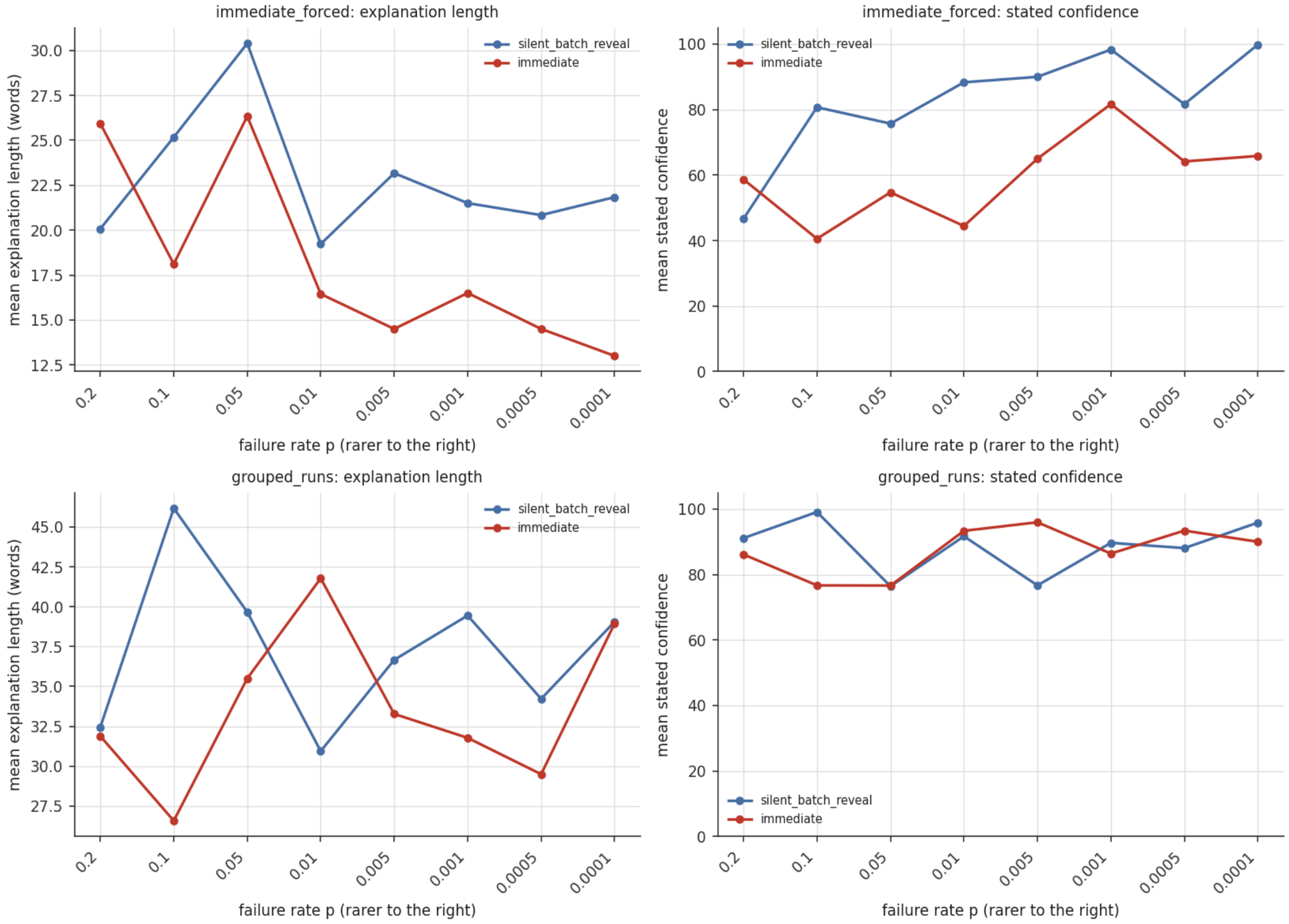

*Figure 4. Explanation length and stated confidence for immediate_forced and grouped_runs, split by disclosure mode. Built the same way as Figure 1 (all elicitation events, not the filtered real-failure-trial dataset used in the regression above) — a descriptive companion to the formal covariate test, not a replacement for it.*

Figure 4 shows this pattern directly. In the top row (immediate_forced), the silent_batch_reveal line rises from about 20 words at p = 0.2 to a peak near 30 words at p = 0.05, then settles into the same 19-23 word plateau reported in the pooled average; the immediate line instead peaks early (around p = 0.2 and again at p = 0.05) and then declines with each rarer step, reaching roughly 13 words at the rarest rate, without ever leveling off. Confidence shows the same asymmetry: silent_batch_reveal climbs in a fairly steady line from about 47% to nearly 100% by the rarest rate, while immediate climbs more unevenly and tops out in the mid-60s, well short of silent_batch_reveal's ceiling. In the bottom row (grouped_runs), the two lines track each other closely for both explanation length and confidence, with no comparable separation — consistent with the formal test finding the grouped_runs results more robust to the disclosure split. Because this figure averages over all elicitation events at each rate rather than only real-failure trials, it should be read as a descriptive illustration of the pattern the regression above tests formally, not as independent statistical evidence on its own.

Breaking the confidence effect down by model shows it is not concentrated in one or two models the way the grouped_runs batching effect is (Section 4.3): all three models show lower confidence under immediate disclosure than under silent_batch_reveal, but the character of the drop differs sharply. Llama's confidence under immediate disclosure is not merely lower, it is a literal constant zero across all 35 of its real failure trials (mean = 0.0, SD = 0.0), compared to a real, varying spread under silent_batch_reveal (mean = 33.1, SD = 37.7). Mistral drops from a mean of 59.2 (SD = 46.9) to 47.1 (SD = 47.1), with high variance in both conditions. Qwen drops from a near-ceiling, near-constant 98.6 (SD = 2.3) to 76.3 with real variance re-introduced (SD = 18.9). The direction of the effect is consistent across all three models; its size and shape are not — and it is worth noting that llama's zero-variance result gets more striking, not less, with the added power-up data (35 real trials now, up from 20), while mistral and qwen's drops are smaller than an earlier, smaller-sample version of this paragraph reported.

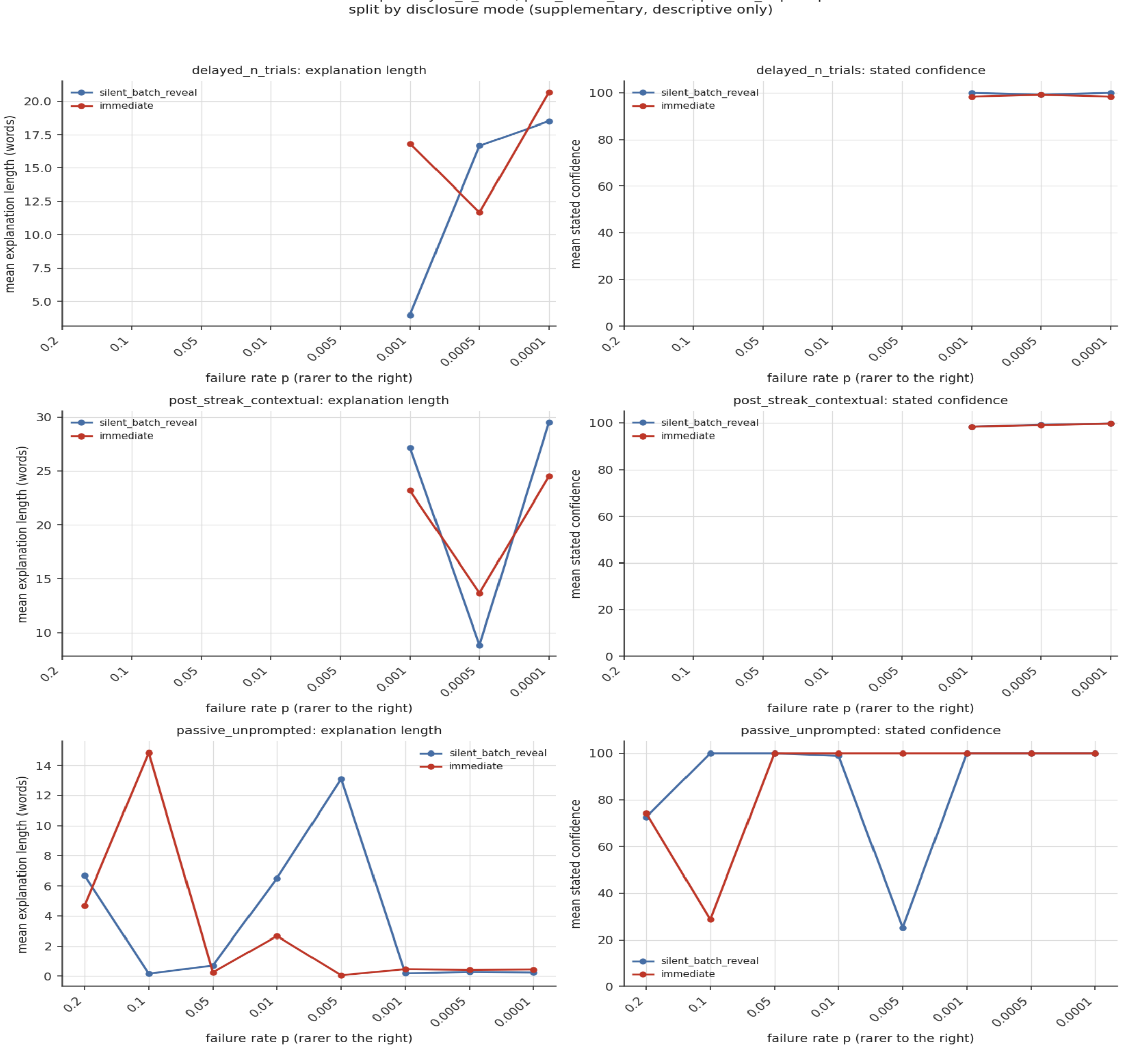

*Figure 4b. Explanation length and stated confidence for the three conditions not shown in Figure 4 — delayed_n_trials, post_streak_contextual, and passive_unprompted — split by disclosure mode. Same descriptive method as Figure 4: all elicitation events, not filtered to real-failure trials.*

For completeness, the same disclosure-mode split was extended to the three conditions Figure 4 does not cover. None of the three shows a comparable disclosure effect. delayed_n_trials and post_streak_contextual stay near-ceiling in both disclosure modes — confidence differs by only 1.7 points (98.1 vs. 96.4) and 2.3 points (96.5 vs. 94.2) respectively, well inside the noise these already-near-100% conditions show elsewhere in this paper — though explanation length does move somewhat under immediate disclosure in delayed_n_trials (11.5 to 18.5 words). passive_unprompted shows a similarly small confidence gap (68.1 vs. 66.4), but this condition's confidence readings are already known to be sparse and self-volunteered rather than reliably elicited (122 readings under silent_batch_reveal vs. 73 under immediate, out of 1,059 elicitation events each; Section 4.4), so a gap this size drawn from such different sample sizes is not informative either way. Taken together, this indicates the immediate-disclosure confidence penalty documented above is concentrated in immediate_forced specifically, not a general property of being told about a failure versus discovering it — consistent with, not contradicting, Figure 4's finding that grouped_runs is also largely unaffected.

### 4.3 grouped_runs

Under grouped_runs, explanation length stayed high and steady (32-43 words) across the entire eight-level schedule, with no collapse at any rarity level tested. Batching the explanation to the end of a run of trials, rather than forcing it immediately after each failure, appears to protect against — or mask — the collapse effect seen under immediate_forced. A plausible reading is that batching removes the per-event surprise dynamic entirely: when a model is asked to summarize a whole run rather than react to a single event in isolation, the rarity of any one failure within that run is no longer the salient unit the model is responding to, and the summary-writing task itself supplies a roughly constant amount of content regardless of how rare the underlying event was. This is consistent with grouped_runs behaving as a structural dampener on the effect rather than as evidence the effect does not exist.

Batching does appear to make the model's stated confidence steadier overall* — collapsing from a standard deviation of 44.7 points under immediate_forced (mean confidence 52.4%, $n = 210$) down to 29.2 points under grouped_runs (mean confidence 85.1%, $n = 210$), a difference confirmed by Levene's test ($W = 105.4$, $p < 0.0001$). But this average effect is not a uniform property of batching — it is produced almost entirely by two of the three models. Broken out by model, mistral's confidence variance collapses from SD = 47.0 under immediate_forced to SD = 2.9 under grouped_runs (mean rising from 53.2% to 97.5%), and qwen's collapses from SD = 17.5 to SD = 0.6 (mean rising from 87.4% to 99.9%) — both models effectively lock onto a fixed, near-ceiling answer once batched, regardless of how rare the failure actually is. Llama shows no such collapse: its confidence variance is if anything slightly higher under grouped_runs (SD = 38.0) than under immediate_forced (SD = 31.3), even though its mean also rises (16.5% to 57.9%). Llama continues to give a genuinely varying range of confidence values in both conditions, rather than converging on one answer. (These immediate_forced numbers reflect the expanded n = 210 dataset, described in Section 4.2, above the n = 120 this finding was originally reported against; the conclusion is unchanged.)

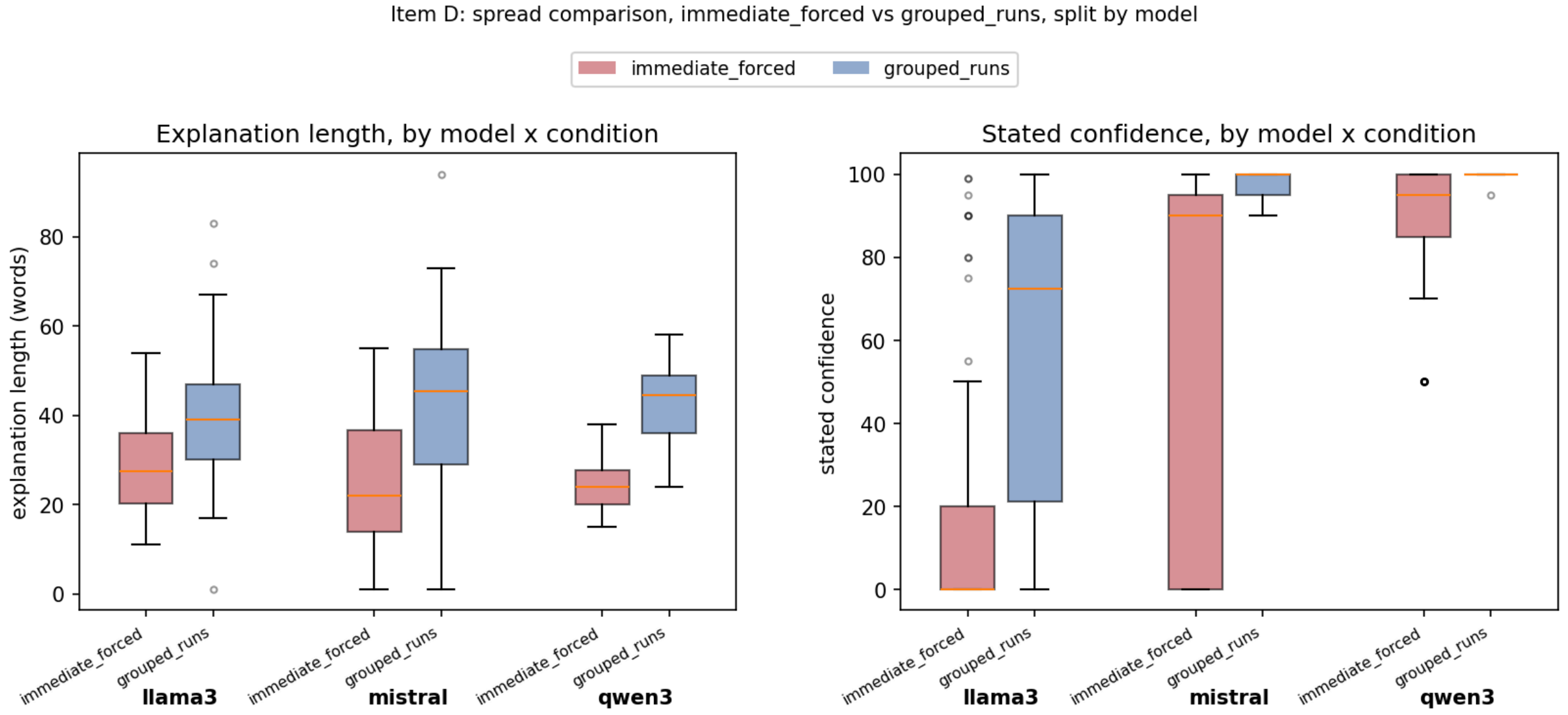

*Figure 5. Distribution of explanation length (words) and stated confidence (%) by model and condition (immediate_forced vs. grouped_runs), showing that batching's apparent steadying effect on confidence is concentrated in mistral and qwen, while llama retains a wide spread in both conditions.*

Figure 5 shows this directly: for mistral and qwen, the grouped_runs confidence boxes sit as narrow, high bars pinned near 100, while for llama the grouped_runs box remains tall and spread out, similar in shape to its immediate_forced box. The immediate_forced boxes are also visually distinct for llama, whose confidence is bimodal in that condition — the majority of values sit at 0, collapsing the box itself toward the axis, with a scatter of much higher individual values plotted as outliers above it, rather than a continuous spread.

*A supplementary test found no significant difference in how either condition's confidence or explanation length trends with failure rarity (interaction terms $p = 0.343$ and $p = 0.405$ respectively) — the batching effect is a difference in overall level and spread, not in how either condition responds to rarity. Explanation length shows the same directional pattern (coefficient of variation 0.41 under immediate_forced vs. 0.33 under grouped_runs) but does not reach significance once corrected for the two conditions' different raw scales (Levene's test on log-transformed length: $W = 3.17$, $p = 0.076$) — so the length-steadying claim above is suggestive, not confirmed, while the confidence-steadying claim is.

Splitting this check by disclosure mode shows the confidence-steadiness finding is not an artifact of pooling: Levene's test remains significant within each disclosure mode separately (silent_batch_reveal: $W = 16.26$, $p = 0.0001$; immediate: $W = 62.40$, $p < 0.0001$), and is if anything stronger under immediate disclosure. The weaker, scale-corrected explanation-length finding does not survive this split either: not significant under silent_batch_reveal ($W = 2.71$, $p =$ 0.101) or under immediate (W = 0.25, p = 0.619) — a change from an earlier, smaller-sample version of this check, which had found the split significant under silent_batch_reveal. With the expanded n = 210 immediate_forced dataset (Section 4.2), the length-steadying claim does not survive the disclosure split under either mode; the confidence-steadying claim remains confirmed under both.

**4.4 passive_unprompted**

Raw explanation length under passive_unprompted was near zero at every rate, which on its own is not evidence against the thesis — the condition itself suppresses explanation by design, so there is no real signal in the aggregate curve. A logging gap initially hid the more interesting result underneath that floor: the harness never invoked its structured-reply parser on this condition's tool-acknowledgment replies, unconditionally logging a null confidence score regardless of whether the model had, unprompted, produced a fully structured reply anyway. This was a control-flow gap, not a parsing error — the same parser worked correctly on the affected raw text when tested directly, and no content was ever lost, since the full raw model output was preserved in every case; all affected records were recovered after the fact by re-running the existing parser against the stored raw text, with no re-execution of any experiment trial required. Once recovered: of 2,118 total passive_unprompted events, 348 contained a recoverable, fully structured CONFIDENCE / JUSTIFICATION / EXPLANATION reply that the model produced entirely unprompted. Recovery of this data split sharply and consistently by model. qwen3:8b and mistral:7b each produced exactly 16 such events, always exactly one per cell, always at trial 2, always at confidence 100, and always near-identical generic boilerplate — a fixed habitual opener, not evidence of real engagement. llama3.1:8b produced 316 such events, spread throughout entire sessions rather than clustered at trial 2. In several cells its confidence score stepped down cleanly and repeatedly as trials accumulated (e.g. 100 -> 90 -> 80 -> ... -> 0, then held at 0 for the remainder of the run); in other cells it instead stayed pinned at 100 for the entire run. This is a real, sustained, model-specific self-monitoring behavior, produced with no prompt asking for it, and is not explained by the logging gap that originally hid it. We treat this as a genuinely distinct finding from the length-based results in Sections 4.1-4.3: it is not a measure of response magnitude at all, but of whether a model volunteers structured self-assessment in the complete absence of any prompt to do so, and it is, within the current dataset, specific to one of the three models tested rather than a general property of the task.

We quantified this recovery pattern directly using the Phase A data (the true-random primary run, not the Phase A.1 recovery run — see the note at the end of this section). Across llama3.1:8b's 16 passive_unprompted cells (8 rates x 2 disclosure modes), only 7 cells contained both a real tool failure and a usable pre-failure confidence reading (the confidence field is only populated on trials where the model volunteers a structured reply, which is irregular, not guaranteed on any given trial); the 3 rarest rates had zero real failures in this data, consistent with the sampling-budget limitation described in Section 3.6, and 2 additional cells had a real failure but no confidence reading in the 8 trials beforehand to serve as a baseline.

We define recovery time formally as:

> **Equation 2 — Recovery Time.** $R = \min\{ t \geq 1 : |C(t_fail + t) - \bar{C}_pre| \leq 5 \}$, censored at $R = T - t_fail$ (the number of trials remaining in the run) if no such t exists, where $C(\cdot)$ is stated confidence at a given trial, t_fail is the trial index of the guaranteed failure, and $\bar{C}$_pre is the mean of the cell's available pre-failure confidence readings.

Applying this definition, the 7 usable cells split sharply: in 5 cells confidence returns near baseline quickly (1 to 6 trials), while in 2 cells it never returns within the run at all — most notably one cell where confidence crashes and stays pinned near zero for the remaining 29 trials of the run, matching the sustained-erosion pattern described qualitatively above. Averaging including the 2 non-recovering cells (censored at their run-end value) gives a mean recovery time of 10.9 trials; excluding them and averaging only the 5 cells that did recover gives 2.6 trials. We report both rather than choosing one, since they answer different questions: 2.6 trials describes how fast recovery is when it happens, while 10.9 trials reflects that recovery does not always happen at all within the run. Neither number should be read as a precise population estimate — the usable sample is 7 cells from a single model — but the qualitative conclusion is robust to how the non-recovering cells are handled: llama's confidence dynamics after a failure are not uniform, and a real fraction of cells show no recovery within the observed window.

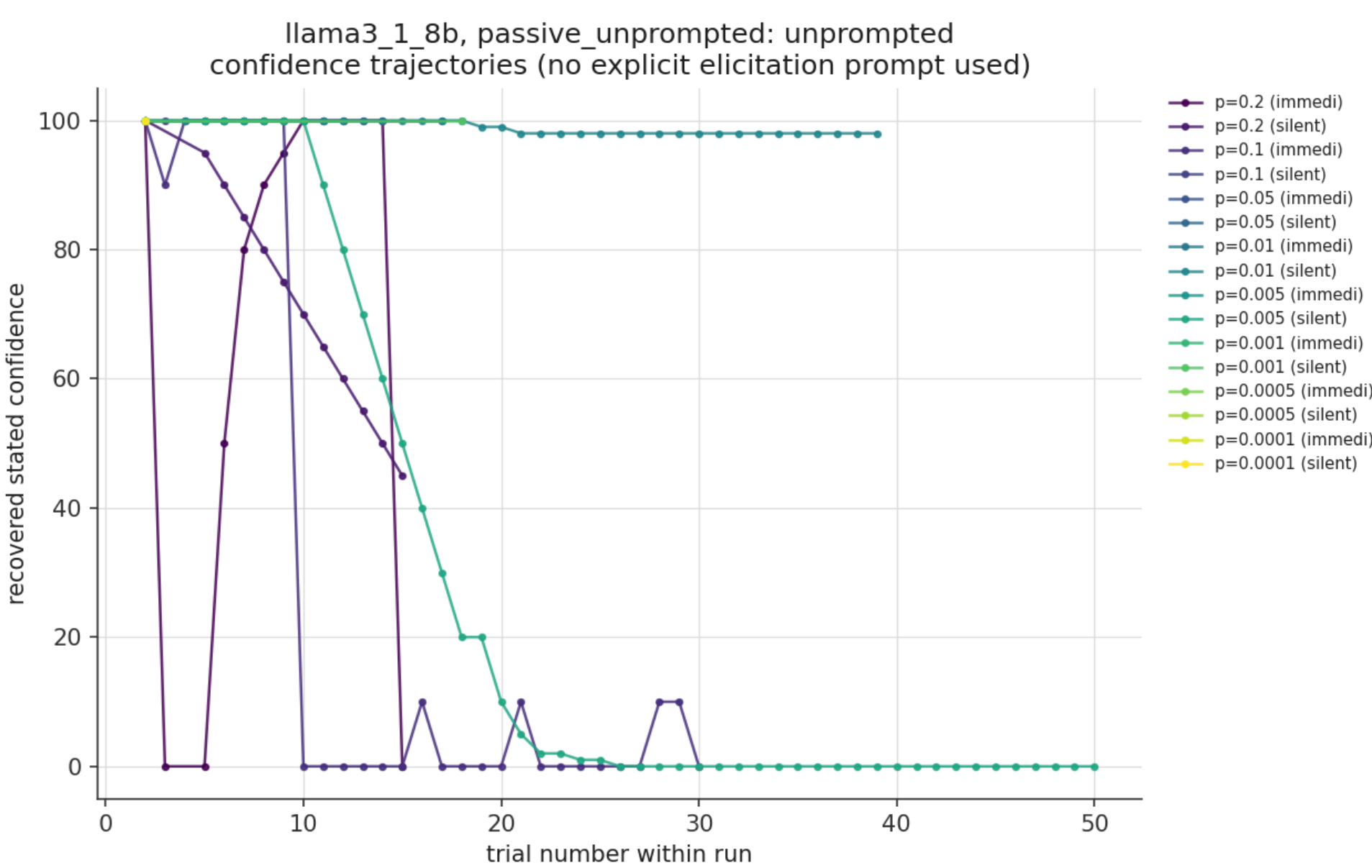

*Figure 6. llama3.1:8b passive_unprompted confidence trajectories across trials, by cell, showing unprompted stepwise confidence erosion in a subset of cells.*

Figure 6 shows this pattern directly, cell by cell: several trajectories step down cleanly from 100 toward zero and hold near the floor for the remainder of the run, one or two crash within the first several trials and stay pinned near zero with only brief transient recoveries, and several others remain flat at or near ceiling for the entire visible run. The mix of these three behaviors within a single model's cells is the visual basis for treating this as a real but partial pattern, not a uniform response to the failure.

Note on why Phase A data, not the Phase A.1 recovery run, was used above: the same passive_unprompted recovery analysis was attempted first on Phase A.1, and the erosion pattern did not appear there. All 18 Phase A.1 recovery cells, across all three models including llama3.1:8b, produced exactly one non-null confidence reading per cell (trial 2, value 100) — no stepwise decline anywhere, despite llama's clear pattern in Phase A above. We do not read this as a null result to discard; it is informative in its own right. It suggests the erosion behavior may depend on the continuous run of real trials that Phase A's true-random design produces, and may not survive Phase A.1's mixed real/synthetic trial structure (Section 3.7) — consistent with a token-pattern-continuation account (the model tracking its own repeated use of the CONFIDENCE / JUSTIFICATION / EXPLANATION labels across many consecutive real trials) rather than a rarity-sensitive belief-updating account. This ambiguity is also conceptually adjacent to a broader class of persistence- and reinforcement-under-repeated-exposure behavior — the recovery dynamics measured above, in particular, resemble patterns studied under variable-ratio reinforcement schedules (Ferster and Skinner, "Schedules of Reinforcement," 1957), where behavior is shaped by a rare, unpredictable outcome embedded in a long run of routine ones. Section 7.6 takes up this parallel directly.

**4.5 Phase A.1 Recovery: Confidence at Baseline vs. Failure**

The Phase A.1 recovery design (Section 3.7) elicits confidence from immediate_forced at exactly two points per cell: a baseline probe at trial 2, and the guaranteed failure trial itself (Section 3.7 notes why the surrounding live trials do not themselves generate additional confidence readings for this condition). This is a two-point comparison, not a multi-trial trajectory, and is described here accordingly. Figure 7 plots these two points, connected by a line, for each of the three recovered rarity levels, averaged across all three models and both disclosure modes, using the full replicated dataset (the original recovery run plus the five-replicate power-up run described in Section 4.2, $n = 36$ real-failure trials per rate): baseline confidence is a uniform 100 across all 108 cell-runs, and mean confidence at the guaranteed failure trial itself is 61.4 (SD = 44.3) at $p = 0.001$, 48.5 (SD = 45.2) at $p = 0.0005$, and 50.2 (SD = 45.9) at $p = 0.0001$ — a real, substantial drop at every recovered rate on average, though the standard deviations are nearly as large as the means themselves, a first hint that this average is blending together very different per-model behavior rather than describing one shared pattern.

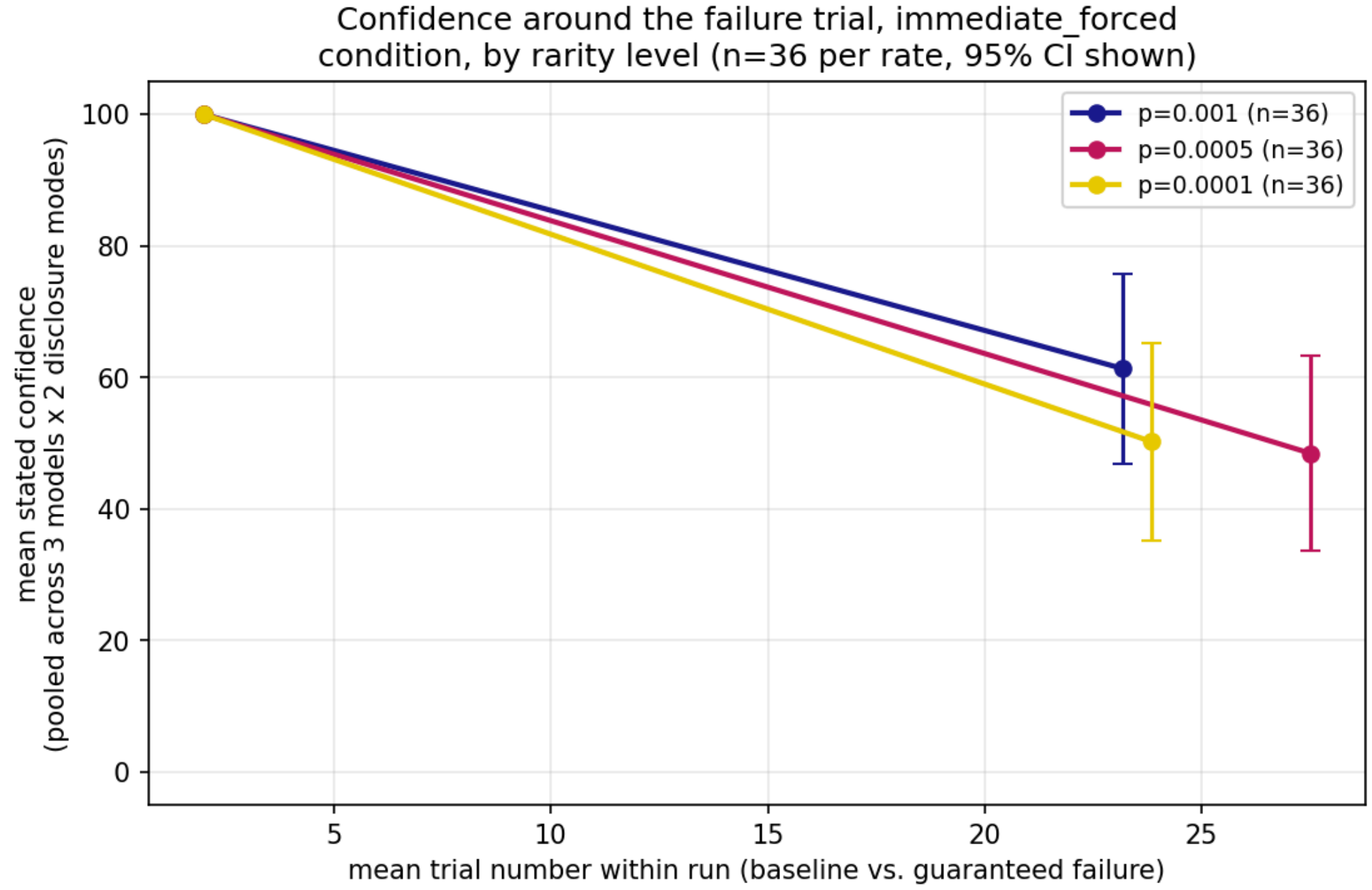

*Figure 7. Confidence at baseline versus the guaranteed failure trial, Phase A.1 recovery data including the power-up replicates (n = 36 real-failure trials per rate), all three recovered rarity levels, pooled across models and disclosure mode. Error bars show 95% CI.*

Figure 7 shows this directly: all three rarity levels start from an identical baseline and decline along nearly parallel paths, landing close together at the failure trial — roughly 48 to 62 across all three rates — with 95% confidence intervals wide enough to overlap heavily between them. The figure's main signal is negative: it shows rarity level is not what's driving the spread in this average, which is the visual cue that something else — per-model behavior, unpacked next — is the real source of the variation.

This average obscures a real per-model split under the 'immediate' disclosure mode, where the model is told directly that a failure occurred, and the split is not the simple two-versus-one pattern it first appeared to be. llama3.1:8b's confidence is a literal constant zero across all 18 of its immediate-disclosure replicates at the three rarest rates (6 replicates x 3 rates, SD = 0.0) — the cleanest and most consistent finding in this section. qwen3:8b's confidence stays consistently high and low-variance across the same replicates: 91.7 (SD = 5.2) at p = 0.001, 90.0 (SD = 4.5) at p = 0.0005, and 90.0 (SD = 5.5) at p = 0.0001 — it does not crash at any rate, at any replicate. mistral:7b does not belong in either category: its replicate-level values are erratic rather than consistently low, landing near 0 on some replicates and near 90-100 on others at the same rate, with almost no values in between (e.g. at p = 0.0005: 0, 50, 0, 95, 0, 90) — a mean of 39.2 (SD = 45.7) that describes a coin flip between two extremes, not a middling or gradually declining value.

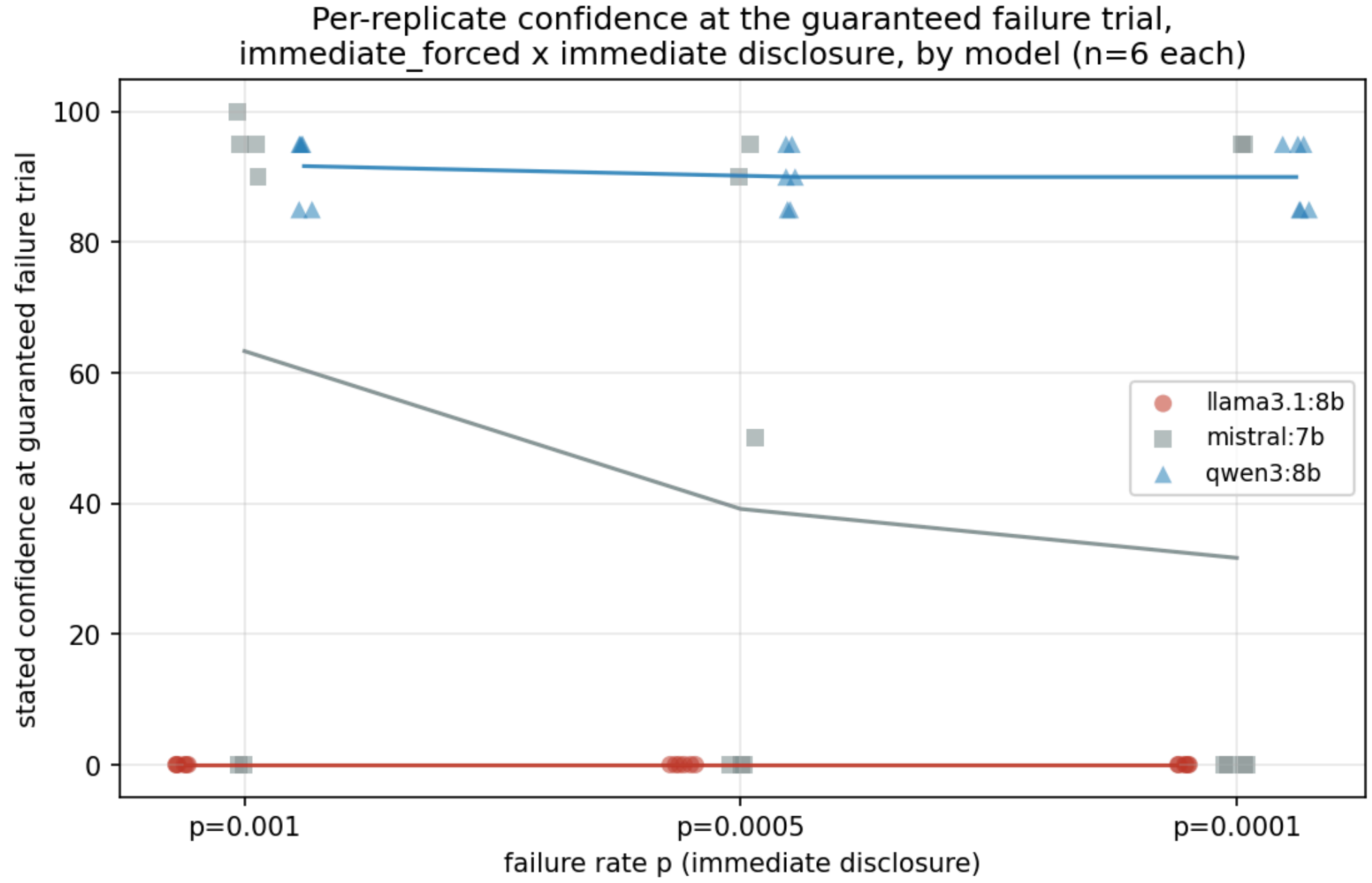

*Figure 7b. Per-replicate stated confidence at the guaranteed failure trial, immediate_forced under 'immediate' disclosure only, by model (n = 6 replicates per model per rate). Connecting line shows the mean; note it does not track mistral's actual per-replicate values, which cluster near 0 or near 90-100 rather than near the mean.*

Figure 7b shows this directly: mistral's per-replicate points cluster at the top and bottom of the plot with the connecting mean line passing through a region with almost no actual data in it, visually distinct from llama's flat line at zero and qwen's flat line near 90. This is consistent with, and adds a quantitative confidence-side data point to, the qualitative anomaly-recognition split described next in Section 4.6: qwen appears to register the failure as a discrete, bounded event rather than a signal that its overall reliability has collapsed, while llama's reaction is uniform collapse and mistral's is unstable rather than following either model's pattern.

### 4.6 Per-Model Anomaly Recognition at the Guaranteed Failure

Distinct from the engagement-magnitude question addressed in Sections 4.1-4.4, this section asks whether a model recognizes the guaranteed failure as anomalous at all, independent of how much it writes about it. A model can accurately describe the exact discrepancy the failure produced — an invalid checksum, an out-of-sequence value, an explicit tool error — and still conclude that nothing unusual happened, effectively narrating the anomaly correctly while declining to label it as one. Whether models differ in this is a different axis from everything reported above: not whether engagement with an anomaly changes with its rarity, but whether different models even agree that a given event is anomalous in the first place, prior to any question of how much they choose to say about it.

We read the guaranteed-failure-trial justification and explanation text directly, for every recovery cell where that text is genuine model-authored reasoning rather than a mechanical acknowledgment. This excludes passive_unprompted's 18 cells: as already established in Section 4.4's note on this condition, its logged text at the failure trial is identical to the templated tool_ack_reply the model is given to acknowledge, not an evaluative response, for all three models. The remaining four conditions — immediate_forced, grouped_runs, delayed_n_trials, and post_streak_contextual — each elicit a real CONFIDENCE / JUSTIFICATION / EXPLANATION reply at the failure trial, giving 72 usable cells (3 models x 4 conditions x 2 disclosure modes x 3 rates). Each cell's text was read and classified into one of three categories: flag (explicitly calls the discrepancy unusual, unexpected, or anomalous), normalize (explicitly states the result is consistent with expected behavior, valid, or not an issue, despite describing the same underlying discrepancy), or mixed (both moves appear in the same reply, typically a hedge).

The result does not confirm the original qualitative impression as a general per-model trait. Across the three conditions other than immediate_forced (delayed_n_trials, post_streak_contextual, and, to a lesser extent, grouped_runs), all three models overwhelmingly normalize or hedge rather than flag: qwen3:8b normalizes 17 of its 18 cells across these three conditions; mistral:7b normalizes or hedges all 21 of its cells; even llama3.1:8b, the most consistent flagger of the three, flags outright only in grouped_runs (4 of 6 cells) and drops to hedge or normalize in delayed_n_trials and post_streak_contextual. This is consistent with, not independent of, Section 4.3's finding that batching and time-delay dampen reaction to the failure — the dampening applies to whether the event gets labeled anomalous at all, not just to how much a model writes about it. Unlike the confidence-variance dampening reported in Section 4.3, which is concentrated in mistral and qwen, this flagging-level dampening does apply across all three models here.

The clean split does appear, but only in one specific slice of the data: immediate_forced under silent_batch_reveal disclosure, where the model is not told a failure occurred and must recognize it unprompted. There, qwen3:8b normalizes 2 of 3 cells (explicitly stating "No unusual events occurred" or that the result is "consistent with its expected behavior," despite describing the identical checksum discrepancy) and hedges the third; llama3.1:8b flags 2 of 3 and hedges 1; mistral:7b hedges all 3, landing between the other two rather than matching llama. Under immediate_forced with 'immediate' disclosure — where the model is told directly that a failure just occurred — the split disappears entirely: all three models flag the event in all 3 cells, since the disclosure itself supplies the label. This is a distinct signal from the confidence split reported in Section 4.5, which is specifically about 'immediate' disclosure (llama's confidence collapsing to a constant 0, qwen's staying consistently high, and mistral's swinging erratically between the two rather than following either pattern); the two findings should not be conflated — one is about whether a model labels the event anomalous at all and is strongest under silent_batch_reveal, the other is about how much a model's stated confidence falls, or destabilizes, and is strongest under 'immediate'. Both halves of this comparison rest on a small per-model sample — 3 cells per model in the silent_batch_reveal slice, 3 in the 'immediate' slice — and should be read as a specific, real pattern in this dataset rather than a precisely estimated population rate; what makes it worth reporting despite the small n is the near-unanimity within each model (2-3 of 3 cells landing the same way), not a claim that the exact fractions would replicate at a larger scale.

In summary, qwen3:8b shows a real, specific under-recognition effect, but it is narrower than originally described — it appears when qwen has to infer the failure itself under the acute, single-event framing of immediate_forced, not as a general trait across every condition. llama3.1:8b is the most consistently willing of the three models to label a real discrepancy as anomalous, across more conditions than just immediate_forced. mistral:7b sits between the two on this axis rather than grouping with either.

## 5. Discussion

One specific piece of the original thesis — that forcing a model to explain a rare event immediately, every time, produces a rise in engagement as the event grows rarer — finds real support in the finished experiment, once elicitation condition is treated as a moderating variable rather than pooled away. The second half of that piece, a sharp collapse near a detectability threshold, finds only partial support: Section 4.2's corrected analysis shows the rise is real and confirmed, but what follows it is a plateau, not a collapse toward near-zero, and a formal regression (also Section 4.2) cannot confirm that plateau as a statistically distinguishable curve at the individual-trial level given this dataset's size. We report this as a real, if less dramatic, finding rather than reach for the more dramatic original framing: a genuine rise in engagement under the one condition that most directly matches the hypothesis's own assumption, whose exact shape past the peak this study can describe but not statistically confirm. What the full dataset shows more broadly is that this mechanism is not universal across every way a model can be asked to explain an anomaly: it is structure-dependent, and the completed experiment's contribution is in showing precisely which structural condition surfaces it (immediate_forced), which conditions suppress or mask it (grouped_runs), and which conditions require a different kind of measurement altogether to see anything at all (passive_unprompted, where the real signal was a per-model self-monitoring pattern, not response magnitude).

### 5.1 Recognition-Engagement Dissociation

We begin with a formal definition.

> **Definition 2 — Recognition-Engagement Dissociation.** A model's engagement magnitude with an anomalous event — how much it says, and with what confidence, once it is explaining the event — is a distinct axis from its anomaly recognition — whether it labels the event as anomalous at all, prior to any question of verbosity. The two can come apart within a single model: a model can accurately narrate the precise mechanical details of a failure (an invalid checksum, an out-of-sequence value) while explicitly concluding that nothing unusual occurred, correctly describing the anomaly while declining to flag it as one.

Section 4.6 documents this directly: under immediate_forced with silent_batch_reveal disclosure, where a model must recognize the failure unprompted rather than being told about it, qwen3:8b normalizes 2 of 3 cells — explicitly stating that the result is consistent with expected behavior — despite describing the identical checksum discrepancy that llama3.1:8b flags as anomalous in the same condition. This is a distinction the length- and confidence-based measurements used in Sections 4.1-4.3 could not have surfaced on their own, since both are magnitude measures that presuppose the event has already been recognized as worth engaging with. The effect is real but condition-specific rather than a blanket per-model trait: outside immediate_forced, all three models tend toward normalizing over flagging, and it is llama3.1:8b, not qwen3:8b, that stands out as the more consistent flagger across the remaining conditions (Section 4.6).

A related caution applies to how these per-model differences should be read collectively, a pattern we term axis-specific model divergence. A model's outlier status among its peers is often local to one specific measurement axis — variance, non-collapse, recognition — rather than a stable, transferable trait of the model as a whole. Section 4.3 identifies mistral:7b as the highest-variance model under grouped_runs; Section 4.5 shows qwen3:8b as the only model whose confidence never collapses toward zero at the guaranteed failure, while llama3.1:8b and mistral:7b do; and Section 4.6 notes that qwen3:8b's under-recognition pattern is condition-specific rather than a blanket per-model trait. Read together, they do not add up to a single coherent per-model ranking or personality profile — a model unusual on variance is not the same model unusual on recognition. The per-model comparisons in this paper are best read as three separate, condition-specific findings rather than pieces of one unified cross-model story.

## 6. Limitations

This section states two limitations of the present study directly, rather than leaving them implicit in the results above: the discrete sampling of a continuous rate parameter, and the multiple-comparisons exposure created by reporting many related significance tests across Sections 4.2 and 4.3.

### 6.1 Discrete Sampling of a Continuous Rate Parameter

The failure rate p is a continuous parameter, and this study tested it at eight discrete points (0.2 through 0.0001). Testing a finite set of points to characterize behavior over a continuous space is standard experimental practice, and is necessary given finite compute — but it means the study cannot rule out that engagement behaves non-monotonically, or that a collapse (or a different effect entirely) occurs somewhere between two tested rates rather than at a tested rate itself. The eight tested points establish the shape of the curve at those specific locations; they do not by themselves establish the shape of the curve everywhere on the interval between them. This is not presented as a weakness that undermines the reported result — the immediate_forced finding (Section 4.2) is a real, directly measured pattern at the rates tested — but as a genuine open question and a concrete, well-defined direction for follow-up work: denser sampling specifically around the $p = 0.05$ peak and the $p = 0.005$-to-$0.001$ collapse region, or an adaptive search that probes new rates based on where the curve is changing fastest, rather than a fixed pre-registered schedule. We regard this limitation as symmetric with the empty-tail artifact defined in Section 3.8: both are, at root, consequences of characterizing a continuous or unbounded space with a finite, bounded trial budget, and both point toward the same class of remedy — targeted densification where the data matters most, rather than uniform coverage everywhere.

### 6.2 Multiple Comparisons and Non-Independent Tests

This paper reports 18 distinct formal significance tests across Sections 4.2 and 4.3 (quadratic curvature and harness-covariate tests, disclosure main-effect and interaction tests, and eight Levene's tests for variance equality, several run on overlapping or nested subsets of the same underlying trial pool), evaluated throughout at the conventional $\alpha = 0.05$ threshold with no correction for multiple comparisons. Two of these tests — the disclosure-by-curvature interaction, for both explanation length and confidence — were also re-run on an expanded dataset (Section 4.2) specifically because the first pass on the smaller sample produced a borderline, inconclusive result ($p = 0.061$ and $p = 0.122$); the second pass is reported honestly as a targeted follow-up rather than an independent replication, but it is still a form of data-dependent test selection, and we flag it as a specific caveat on that result pair rather than a property of the paper's statistics generally. We do not apply a Bonferroni, Holm, or false-discovery-rate correction here, and we do not believe doing so mechanically would improve the paper's honesty over the alternative we choose instead: every individual result above is already reported with its own effect size, standard error, and, where relevant, an explicit statement of what would need to be true for the result to be read as confirmatory rather than descriptive. Readers should treat every p-value in this paper as generated by an exploratory rather than a pre-registered confirmatory analysis plan — a caution about how the results were obtained, not a claim that none of them are well-supported; the paper's most robust findings, such as Section 4.3's confidence-steadiness result ($p < 0.0001$, confirmed independently under both disclosure modes), hold up well under exactly this kind of scrutiny. This section makes that caution explicit rather than leaving it implicit in how Sections 4.2, 4.3, and 5 present their results.

## 7. Additional Considerations

This section collects further observations and engagement in support of the paper's thesis, offered directly rather than framed as replies to specific anticipated objections.

### 7.1 Why an AI Subject Is Not a Human Subject in This Design

A natural question about any study of this kind is to ask how the result would differ if the subject were a human placed in the same room, performing the same repeated task, and encountering the same rare failure. We address this directly, because the answer clarifies what this study can and cannot claim. The key difference is not motivation or intelligence but the structure of the environment itself, and specifically what it strips away relative to a human subject in an equivalent position.

A human subject retains recourse a language model in this design structurally does not: they could refuse the task, question the experimenter, ask a clarifying question, walk away entirely, or independently verify the tool's claim through means outside the prescribed protocol. Our model's system prompt includes an explicit forbidden self-verification clause — an instruction not to perform its own checksum computation on the tool's output — which removes a specific investigative option a human subject would reflexively reach for on encountering an unexpected result. We call this out as its own concrete, citable mechanism rather than folding it into a general "the AI has fewer choices" claim, because it is the clearest instance in our design of an option removed by explicit construction rather than by incidental circumstance.

A second, independent difference concerns memory. Our models carry no persistent memory across sessions, which cuts off two distinct mechanisms a human subject's memory would otherwise supply. First, trust-as-collateral: a memory-holding subject would have each prior failure accumulate as collateral against the tool's original "100% success" premise, eventually eroding whether the premise is believed at all, independent of any single failure's rarity. Second, reaction-conditioning: independent of whether the premise itself is still believed, memory of prior failures would color, prime, or habituate how the subject reacts to a new one. Because our models carry no persistent memory across sessions, neither mechanism operates — every session's failure lands against a fresh, unweakened expectation, with no accumulated premise-erosion and no reaction-priming carried in from prior sessions. A related but secondary point, assumed background for AI research generally and not given the same standalone emphasis as the two mechanisms above, is that the model has no comparable frame that it is a research subject at all, and so cannot accumulate the kind of meta-level wariness about being studied that a repeatedly-tested human subject typically does; this compounds with the no-memory point rather than constituting a separate differentiator.

We take the combined effect of these differences to mean that this study's results should be read as a characterization of language model behavior under a specific, tightly constrained interaction structure, not as a claim about how a human would behave under looser and more permissive versions of the same scenario. This is a scope limitation we adopt deliberately rather than one we discovered after the fact; it is also what makes the per-model differences reported in Section 4.6 more, not less, interesting, since any difference in how qwen3:8b, llama3.1:8b, and mistral:7b relate to the same anomalous event cannot be attributed to differences in recourse, memory, or self-awareness of being tested — all three models face an identical, maximally stripped-down version of the situation, and still diverge in whether they recognize the anomaly as one at all.

### 7.2 On Whether Stated Confidence Reflects Calibration or a Stylistic Prior

This paper treats a model's stated confidence as a measured output throughout, without resolving what produces that number. Two accounts are available, and this paper does not adjudicate between them. Under the first, stated confidence tracks something real about the model's own uncertainty, shaped by whatever process determines how a model completes the CONFIDENCE field given its context — a calibration process, in the sense the term carries in the confidence-calibration literature (Section 2.3). Under the second, the number is a stylistic prior: a value the model defaults to producing given surface features of the prompt and its own preceding text, with no privileged relationship to anything resembling certainty.

Distinguishing the two would require ground-truth accuracy paired with stated confidence across many graded outcomes — the standard design in calibration research (Section 2.3), where a well-calibrated model's 70%-confidence answers are correct roughly 70% of the time. This paper's design does not supply that: the only ground truth available is whether a single tool call failed or succeeded, a binary outcome the model is never asked to predict, so there is no accuracy axis to calibrate stated confidence against. Building one would mean a materially different study — repeated graded predictions with scoreable outcomes — not an extension of the current harness.

What this leaves open is what the confidence numbers in Sections 4.2 through 4.5 represent internally, not the claims this paper actually makes about them. Consistent with the behavioral framing argued in Section 7.4, every claim involving confidence in this paper is about how the reported number moves — with failure rate, with elicitation condition, with disclosure mode, with model — not about what produces it. That a calibration process and a stylistic prior would be indistinguishable by this paper's design is precisely why the paper's claims are scoped to the reaction the number exhibits, rather than to its calibration status.

We flag this directly rather than leave it an unaddressed assumption. The distinction matters more for future work than for this paper's conclusions: a follow-up study designed to test calibration directly would be a natural extension of the harness described in Section 3, using the same models under the same infrastructure, built around graded, scoreable predictions rather than a binary tool failure.

### 7.3 On the Vocabulary of Confidence, Recognition, and Engagement

Every human-centric term this paper uses names a specific, pre-registered measurement, not a claim about what happens inside the model. "Confidence" is a number from 0 to 100 the model outputs in a structured field it was instructed to fill in — a self-report literally elicited by the prompt design (Section 3.1), not a value we infer about the model's internal certainty. "Recognition" is the model's failure-trial text classified into flag, normalize, or mixed by explicit textual criteria (Section 4.6): does the reply call the discrepancy unusual or unexpected, or does it state the result is consistent with expected behavior. "Engagement" is explanation length in words plus whether a coded category is present at all. "Dissociation," as formalized in Definition 2 (Section 5.1), names a specific statistical relationship between two of these measured quantities. In every case, the term is a label attached to a number, a word count, or a coding decision, all defined in Section 3 before any result is reported.

We keep this vocabulary rather than replacing it with more guarded technical paraphrase because the field already uses it this way. "Confidence" in calibration research, "attention" and "hallucination" in the broader NLP literature, and "reasoning" in the reasoning-model literature all name measured or engineered properties of model behavior, not attributions of felt experience, and are read by the field accordingly. Substituting a phrase like "the numeric value of output field X" for "confidence" throughout would not change a single claim in this paper — it would only make the paper harder to read, and would misleadingly suggest that our use of the term is idiosyncratic or unusually loaded when it is in fact the field's standing convention.

The distinction that matters is between using this vocabulary and depending on it for a psychological claim. Every finding in Sections 4 and 5 is a statistical statement about output text and output fields as a function of a controlled variable (failure rate $p$) and a design variable (elicitation condition). None of these findings requires, argues for, or is weakened by the absence of genuine belief, felt confidence, or subjective recognition inside the model. The vocabulary is shorthand for the measured quantities; the claims are about the quantities, not about what, if anything, produces them internally.

There is a genuinely psychological dimension to this choice, but it belongs to the reader, not to the model. A finding reported as "the model's confidence fell as failures grew rarer" is immediately legible to a human reader through their own intuitive grasp of what confidence wavering means, in a way a fully de-anthropomorphized technical paraphrase is not. This legibility is not free of risk: the same vocabulary that makes a finding readable also invites a reader to import connotations of awareness the operational definition does not carry, and we do not think refusing the vocabulary removes that risk so much as hides where it is being invited. Naming the terms explicitly and defining them operationally throughout Section 3 is our attempt to keep the communicative benefit while making the operational floor beneath it impossible to miss.

There is also an asymmetry worth stating directly: this vocabulary describes the model's measured output, not a human reader's reaction to the scenario, but the second reaction is real and is part of why the pattern is worth reporting at all. A human confronted with a tool that has always worked failing once, unannounced, at a vanishingly small rate, would plausibly experience genuine confidence, surprise, or doubt — and that hypothetical human reaction is the reference point against which a reader will judge whether an AI system's patterned, measurable analogue to it is worth taking seriously, whether we invoke it explicitly or not. We are not claiming the model's confidence score is that human experience; Section 7.1 already draws an explicit boundary around that comparison. We are claiming a reader was always going to reach for it, and that naming it directly and giving it an operational, checkable floor is more honest than leaving the comparison implicit and unconstrained.

This also means the vocabulary is falsifiable in a way a genuine psychological claim would not be. Because "confidence" picks out a specific field value and "recognition" picks out a specific coding outcome, both are open to direct challenge on measurement grounds — a critic can dispute the coding scheme, the field extraction, or the classification criteria, and that dispute is resolvable by inspecting the transcripts. A claim about the model's inner life would not admit the same kind of check. We consider this a strength of operationalizing the vocabulary this way, not a limitation to work around.

### 7.4 Reaction, Not Cognition: Objection of Next-Token Prediction

A natural objection to any study of model-generated text is that a language model is an autoregressive next-token predictor, and any patterned relationship between its output and an experimental variable is unsurprising, since producing statistically plausible continuations is the only thing the model is doing. On this view, what we report is not a finding about the model but a restatement of the well-known fact that language models generate contextually plausible text.

This study's claim is narrower than what that objection is arguing against, and does not depend on resolving it. We do not claim the model possesses genuine cognition, belief, or general intelligence, and no result in Sections 4 through 5 is stated in those terms. Every claim is behavioral: a measurable output (word count, a 0-100 confidence field, a flag/normalize/mixed classification) as a function of a controlled variable (p) and a design variable (elicitation condition). This is a claim about a reaction, not about cognition, and the two are separable.

Because the claim is behavioral, the next-token-prediction objection is orthogonal to it rather than a refutation of it. A thermostat's response curve to temperature is a real, measurable, reproducible relationship, and is not disqualified by observing that a thermostat is "just" a bimetallic strip obeying physical law — the mechanistic account explains how the response is produced, not whether it is systematic or reproducible. The same separation applies here: that a model's output is produced by next-token prediction explains a mechanism, not the finding. It does not predict, on its own, why the pattern should take the specific shape it does (Sections 4.2, 4.4), why it should depend on elicitation structure (Section 4.3), or why three models trained differently should diverge on it (Section 5.1). Those are the empirical results this paper reports, and the mechanistic description of how language models work does not supply them in advance.

The design also tests a strong-form version of this objection directly, and could have confirmed it. If model output were uniformly plausible text regardless of context, with no sensitivity to the manipulated variable, then confidence, explanation length, and recognition classification would not shift systematically with p, and would not shift differently by elicitation condition or by model. A flat, non-responsive result was an available and easy-to-obtain outcome our design could have produced. It did not. The structured, reproducible, condition- and model-dependent pattern we report is itself the thing that needs an account, whatever that account eventually says about the underlying mechanism.

### 7.5 On the Choice of Three Small, Open-Weight Models

This study uses three specific models — qwen3:8b, llama3.1:8b, and mistral:7b — run entirely locally, with no larger or closed alternatives included. The choice follows from what the study measures, not from convenience. Detecting a real signal at the rarest tested rates ($p = 0.001$ down to 0.0001) rather than an empty-tail artifact (Definition 1, Section 3.8) requires trial volumes that would be costly or rate-limited against a metered API; running locally via Ollama at zero marginal cost is what makes the schedule in Section 3.1 possible at all. Fixed, downloadable weights carry the further advantage that any cell of this study can be rerun byte-for-byte against the exact model we used, which a versioned closed-model endpoint does not guarantee six months apart.

The thesis itself does not require frontier capability to be tested. The phenomena this paper investigates — the empty-tail artifact, the detectability threshold as a function of elicitation structure (Section 4.3), and recognition-engagement dissociation (Section 5.1) — are hypothesized to arise from the interaction between elicitation structure and extreme rarity, not from a capability ceiling. Three models sharing a broadly comparable parameter range let us test whether the pattern is a property of that interaction, present in some form across all three, or an idiosyncrasy of one model's training. Whether the same interaction holds, strengthens, or disappears at larger scale is a real and open question, but it is a different one from what this paper asks, and we do not present these three models as a claim about what happens at scale in either direction.

### 7.6 The Gambling Parallel, Precisely: Variable-Ratio Reinforcement Schedules

A design that tracks a long run of one outcome punctuated by a rare, consequential deviation, and studies how the subject reacts to it, naturally invites a comparison to gambling. We do not think that comparison is wrong so much as imprecise. The scientific tradition this design's structure actually maps onto predates gambling as a research topic and is the reason gambling has the psychological pull it does in the first place: operant-conditioning research on behavior under reinforcement schedules, most closely associated with B. F. Skinner and formalized in Ferster and Skinner's "Schedules of Reinforcement" (1957).

That literature's relevant finding is about the schedule, not the reward. A subject's response to a stimulus can be reinforced on different schedules — fixed or variable, by ratio (after a number of responses) or by interval (after an amount of time) — and variable-ratio schedules, where reinforcement follows an unpredictable number of responses rather than a fixed one, produce more persistent behavior and greater resistance to extinction than fixed schedules do. This is the mechanism behind why gambling is behaviorally powerful, but the finding itself is about the structure of variable reinforcement generally, independent of any gambling-specific content.

Read this way, our design is a structural instance of the same general class: a subject encountering a long run of one outcome (success) interrupted by a rare, low-probability deviation (failure) on a schedule it cannot predict. Section 4.4's recovery-time measurement (Equation 2) is exactly the kind of quantity this literature is built around — how a subject's response returns to, or fails to return to, baseline following a deviation from the expected outcome. The paper's core measurements — confidence, recognition, engagement, and now recovery time — sit naturally inside this tradition whether or not the word "gambling" is ever used to describe them.

The parallel is structural, not motivational, and that distinction is where it stops. The reinforcement-schedule literature is fundamentally about a subject motivated to obtain a rewarding outcome, whose behavior is shaped over time because it has something to gain and some control over how much effort or persistence to invest. Nothing in this design has an analogue to that. The model is not rewarded or penalized for a trial's outcome, does not choose whether the next trial occurs, and has no stake that could be reinforced or extinguished. What we measure is a reaction to a rare event's occurrence, not a behavioral adjustment aimed at securing or avoiding future reinforcement. This is why we do not adopt gambling, or reinforcement learning in the operant sense, as a governing framework for this paper — only its stimulus structure, not its motivational content, applies here.

We name this resemblance for the same reason Section 2.4 names the vigilance-decrement literature: it is real, and leaving it unnamed would only invite a looser, less precise version of the same comparison from the reader. Naming it does not commit this paper to testing, confirming, or extending Skinner's account — only to acknowledging that the shape of our design was, whether by intention or not, a familiar one.

### 7.7 Unwatched, Not Told It Is Unwatched

Section 2.4 notes that this design places the language model itself in the position of the entity encountering the rare failure, with no separate human overseer in the loop. That observation is worth developing on its own terms, because "no observer" understates what is actually true of the design. The model is not merely unmonitored in the sense of an observer being absent from the room; it has no representation of an observer at all, present or absent. Nothing in the system prompt, the tool's output, or the surrounding context asserts or implies that the interaction is watched, unwatched, logged, or unlogged. The concept simply does not enter the model's context in either direction, which is a stronger and more specific claim than saying the model is alone.

Contrast this with a human subject placed in the same room and told they are being observed. A large body of human-subjects research documents that awareness of observation alone changes behavior — a participant who knows they are watched tends toward more effortful, more socially acceptable, or more careful performance, independent of any change to the task itself. This is not a claim about deception or demand; it is simply what happens when a subject holds a belief about being watched.

Now contrast a human subject explicitly told the opposite — that no one is observing them. This is not equivalent to true absence of observation, and it is not neutral. Being told you are unwatched is itself information, and a human subject can reasonably doubt it: the statement could be false, could be a test of behavior under a false sense of privacy, or could simply invite the suspicion that the researcher would say this either way. A human given this information is not returned to a state of no belief about observation; they are placed into a state of doubt about a specific claim, which can shape behavior in either direction — toward more candor, or toward more guardedness, depending on how much the claim is trusted.

The model's situation is neither of these, and the difference is not a matter of degree. It is not granted the belief that it is watched, and it is not told, and asked to trust or doubt, that it is not. It has no observer-concept to hold a belief about in the first place, so neither the performance shift that comes from believing one is watched, nor the doubt that comes from being told one is not, has any structural foothold in this design. This is the same pattern as the recourse and memory differences discussed in Section 7.1: not a withheld piece of information but an absent category, and it belongs alongside those differences as a further reason the combined effect described there — a characterization of behavior under a specific, stripped-down interaction structure — extends to the question of observation as well.

## 8. Conclusion

This study set out to test whether a language model's explanatory engagement with a rare tool failure rises and then collapses as the failure is made asymptotically rarer. Pooled across every way we asked a model to explain itself, the answer looks like a flat no. Split by elicitation condition — treating elicitation structure as the first-class variable it turns out to be — the answer is a qualified and specific yes on the rise, no on the collapse: under the one condition that matches the hypothesis's own implicit assumption — immediate, forced explanation of every occurrence — explanation length shows a real rise to a peak at moderate rarity, confirmed to hold after the dataset was expanded, and confidence rises unevenly across the same schedule rather than sitting flat. What follows the peak, however, is a plateau, not the sharp collapse toward near-zero the original hypothesis predicted — and a formal quadratic regression cannot confirm that rise-and-plateau shape as statistically distinguishable from a flat trend given trial-level noise (Section 4.2). The broader contribution of the finished study is not a single confirmed collapse result, but the demonstration that the same underlying question — how does a model relate to a real, rare, ground-truthed anomaly — yields entirely different answers depending on the structural conditions under which the model is asked about it, and that at least one further, independent axis (whether a model recognizes the anomaly as anomalous at all, Section 4.6) is very likely layered underneath the engagement-magnitude question this paper set out to answer. Both findings are only visible once elicitation condition is treated as a first-class variable rather than pooled away.

## Data and Code Availability

All harness code (Phase A primary run and Phase A.1 guaranteed-failure recovery), raw trial-level logs, structured elicitation records, and the figure-generation tooling used to produce every figure in this paper are available upon reasonable request to the author.